# Continuous and large-scale: ELEANOR, the soft architected arm inspired by the elephant trunk

Giovanna A. Naselli[1†], Anderson B. Nardin[1†], Seonggun Joe[1,2†], Ryan Drinkwater[3],

Enrico Donato[4], Diego Bianchi[4], Egidio Falotico[4], Michel C. Milinkovitch[5] and Lucia Beccai[1]*

[1] Soft BioRobotics Perception Laboratory, Istituto Italiano di Tecnologia, 16163 Genova, GE, Italy

[2] Department of Autonomous Systems Engineering, Korea Aerospace University, Seoul, South Korea

[3] Photocentric Ltd., Peterborough PE1 5XN, United Kingdom

[4] BRAIR lab, The BioRobotics Institute, Scuola Superiore Sant'Anna, 56025 Pontedera, Italy

[5] Laboratory of Artificial and Natural Evolution, Department of Genetics and Evolution, University of Geneva, 30, Quai Ernest-Ansermet, 1211 Geneva, Switzerland

[†]G.A.N., A.B.N., and S.J. contributed equally to this work

* Corresponding Author E-mail: lucia.beccai@iit.it

## Abstract

The elephant trunk is a dexterous and versatile manipulator whose performance is still unmatched in robotics. In previous works, modularity was prioritized and relatively small-scale continuum robots were built. We take the natural proboscis of the *Loxodonta africana* species as a model and propose a different design approach which favors structural continuity and natural frequency that plausibly emulate those of the natural trunk, while conferring high adaptability to the environment and humans. Instead of targeting prescribed behaviors, we show that a biomimetic design based on the macroscopic properties of the natural system enables elephant-like movements and grasping. We build by 3D printing an 85 cm long, compliant, tapered, volumetrically tessellated continuum arm, which is combined with tendon-driven actuation mimicking the longitudinal and oblique muscles of the natural model. We demonstrate whole-body grasping of objects having different shapes and dimensions and discuss a comparison to the biological trunk highlighting aspects of both biology and robotics.

## Introduction

The elephant trunk is a large-scale, boneless, extremely versatile organ. Besides serving physiological (*e.g.*, breathing, smelling) and social (*e.g.*, disciplining, offering comfort, communication) functions, it accomplishes a wide spectrum of mechanical tasks: elephants use it to manipulate a diversity of items (*e.g.*, of different dimensions, shape, stiffness, quantity, physical state, etc.) to feed themselves and for self-defense, interacting with their conspecifics and with the environment (*1*, *2*). The fact that this organ does not include a skeleton allows for continuous deformations, as it holds for all the muscular hydrostats (*3*). Its structure is notably complex: an arrangement of dozens of thousands of muscular fascicles (*4*), wrapped in a tough yet deformable skin, together with connective tissue (*5*). Owing to its dexterity and versatility, it is a remarkable model for robotic grasping and manipulation, different from the human arm-hand system (*6*, *7*). Thus, it has inspired a variety of continuum and hyper-redundant arms, consisting of rigid components, soft parts, or both, and recurring to various actuation principles (*e.g.*, tendon-driven, fluidic). The lengths of these robots are in the range 30—85 cm, depending on their structure. About two decades ago, researchers built an ~85 cm long hyper-redundant robotic trunk, consisting of sixteen serially connected modules distributed among four independently actuated cable-driven sections (*8–10*). This robot, teleoperated by a joystick, implemented large displacements to perform whole-body grasping with limited versatility. Since then, multiple robotic solutions have been proposed, like an arm incorporating flexible rods (*11*), a tendon-driven arm which performs grasping through deformation patterns based on spirals (*12*, *13*), robotic modules based on dielectric elastomers (*14*), slender arms based on shape-memory alloys (*15*), rigid-flexible coupling modules (*16*), pneumatic actuators often interconnected by rigid components (*17–20*), and rigid-linked modular robots (*21–24*). Some allow for pre-programmable stiffness via a combination of soft pneumatic chambers and rigid components (*19*) or recur to modular tensegrity structures (*25*, *26*). Other researchers tuned the stiffness of each module of their robot through numerical structural models (*27*), and recently reported how modules made of lattice structures can be assembled in a ~30 cm long arm (*28*).

Yet more efforts are needed to narrow the performance gap between the biological and the artificial systems, especially at the meter scale. Above all, it stands out that there is a notable mismatch between the hardware of existing trunk-inspired robots and biological architecture. Although it is known that the currently available technology poses limitations (*29*, *30*), the problem does not entirely lie in the different complexity between the natural and the artificial trunks. Indeed, it has been found that elephants rely on strategies to reduce that complexity, as their trunk works based on the use of a few, well identified, muscular groups (*31*, *32*), which suggests that hardware simplification is legitimate. Building on those findings, other works (*33*) proved that some of the out-of-plane deformations achievable by the elephant trunk are enabled solely by the relative orientation of those groups, showing that muscular arrangement is one crucial feature in this case. This led to trunk-like structures that can be maneuvered in an effective way through a biomimetic placement of a few actuators (*34*). However, the implementation remains at a relatively small scale (~20 cm). Building a large-scale bioinspired robot presents major challenges that we address here.

For continuum arms, especially if soft, size plays a key role from the structural viewpoint, since a trade-off between mass and stiffness should be found (*35*). One must deal with undesired large static deflection and buckling, not to mention durability, mechanical strength, and possible issues in the fabrication process – important matters especially when envisioning deployment in the real world. A large stiffness vs. mass ratio would mitigate at least some of these issues, but such a design approach would fail the purpose of biomimetics and would lead to systems with high natural frequencies – not a favorable feature for continuum arms intended for close interaction with human operators, and indeed not a characteristic of elephant trunks. Let us note, in fact, that an adult elephant can easily lift sixty kilograms using its trunk (*36*), still the trunk itself is a heavy structure, having a mass of about 100—150 kg (*36*, *37*) which is brought about by a much heavier body. Thus, considerations on its performance should not exclude its self-weight. Then, should a large trunk-like robot weigh as much as a biological trunk? Or, better, should it have the same natural frequencies? So far, most research efforts aim at mimicking the elephant's trunk deformability to achieve predetermined motions. With a different perspective, we commit to a radically different approach: we suggest that artificial implementation should mimic the biological counterpart even for the natural frequency. Indeed, dynamic scaling has already been discussed to tackle the development of biomimetic locomoting robots (*38*). However, to the best of our knowledge, it has not yet been adopted in the field of continuum manipulators.

Adopting this point of view, a major concern is the selection of suitable materials. Previous studies (*39*) estimated that the apparent Young's modulus of the elephant trunk is 938 kPa, and muscular tissue has a reported density of 1180 kg/m$^3$ (*40*). Materials typically used in the field of soft robotics have elastic moduli of the same order of magnitude (~1 -10 MPa), and similar density (~1000 kg/m$^3$). Hence, we might think that large robotic trunks could be built, in principle, with soft materials. However, a bulk artificial trunk made of hyperelastic material would hardly host a functional, biomimetic muscular arrangement. To cope with these aspects, architected structures (like lattice arrangements) offer the possibility not only of tuning stiffness, but also of implementing tendon-driven actuation, since tendons can be routed through the unit cells of the overall structure, as shown in (*28*). In that work, researchers tune the stiffness of a series of modules to preprogram their deformation. In contrast, we privilege continuity over modularity. Our choice is motivated by previous experiments with animals suggesting that mechanical continuity is crucial: the elephant trunk not only bends and twists largely, but it does so with continuous and smooth curvature, and wraps around an external object establishing contact over a large area (*41*).

In this work, we achieve whole-body grasping and elephant-like deformations by implementing a trunk made of lattice and leveraging tapering to tailor the flexural rigidity along the trunk's longitudinal axis. In particular, the main goal is to achieve a large-scale system that can deform as smoothly as its biological counterpart. The design is based on biological information extracted from a *Loxodonta africana* elephant trunk, and the unique large-scale architected structure is entirely built by 3D printing of an *ad-hoc* soft material with a printer having a large build volume. The tendon-driven soft ELEphAnt-iNspired large-scale cOntinuum and continuous aRm shown in Fig. 1, hereafter named ELEANOR, is 85 cm long. We show that dynamic behavior plausibly mimics that of the natural trunk and focus on the whole-body grasping capabilities through minimal biomimetic actuation. Eventually, we discuss the results under both the robotic and the biological standpoint, with the overarching research question being: what is the best approach to build trunk-inspired manipulators?

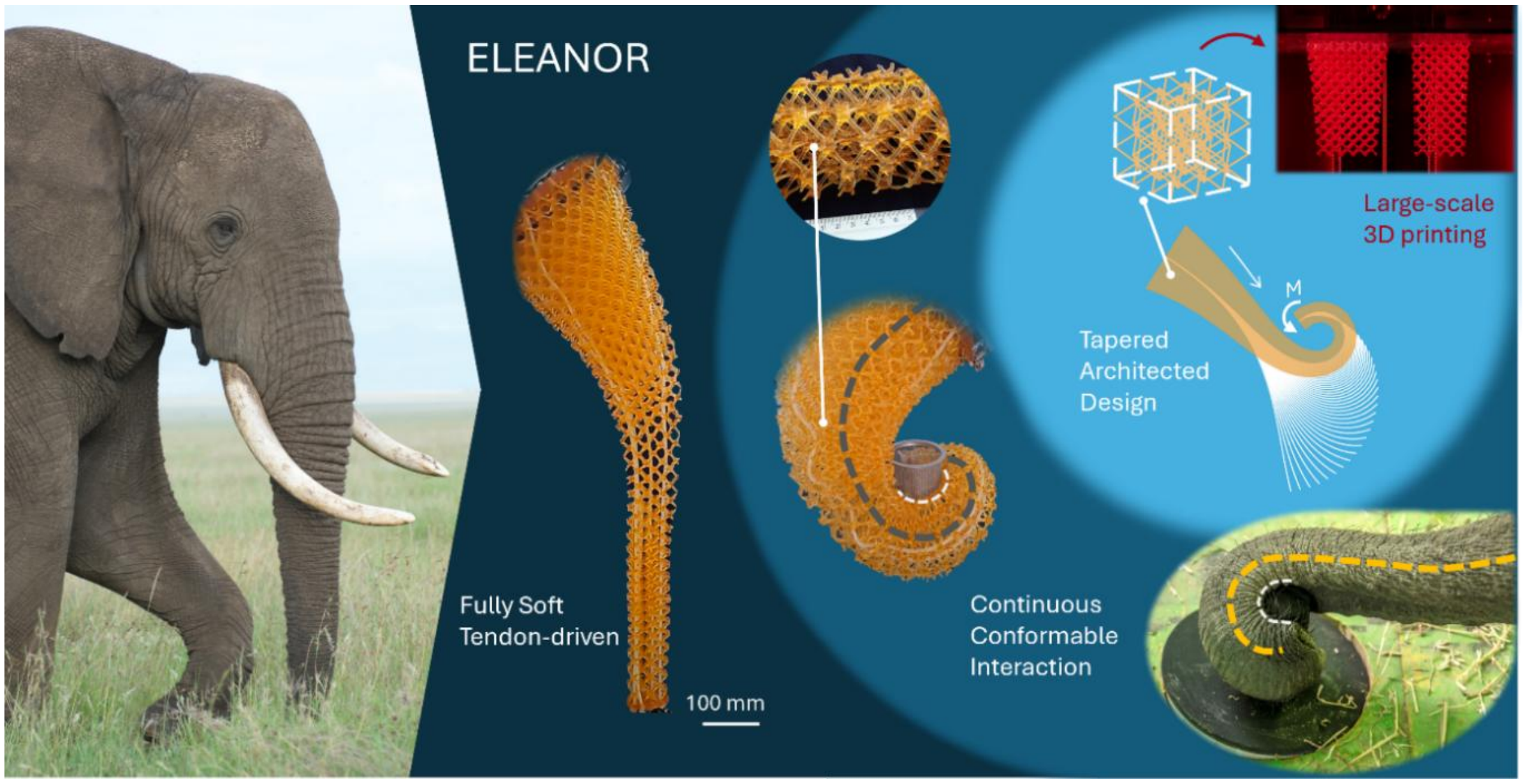

**Fig. 1. Large-scale soft continuum arm inspired by the elephant trunk.** ELEANOR is a large-scale, fully soft tendon-driven robot that leverages structural continuity to achieve elephant-like behavior. It has a tapered body made of architected material, obtained through 3D printing of three large-scale parts that are joined seamlessly. The compliance of its structure enables high conformability to the environment; its overall mechanical properties allow for versatile grasping, as its distal region wraps around the grasped object as performed by elephants. The photograph of the elephant trunk at the bottom right is of Matteo Montenero.

## Results

### *Design based on biological data*

The design of ELEANOR was defined based on the dimensions and the shape of the trunk belonging to an individual of *Loxodonta africana* also used in a previous study and available from an open repository (*32*, *42*). The dimensions of the biological trunk (including length, dimensions at the base and at the tip) are rescaled of a factor ~0.5, accounting for fabrication and actuation-related aspects (see Materials and Methods). The resulting artificial trunk is 850 mm long and has a proximal end cross section of dimensions 255×235 mm (from left to right and from dorsal to ventral, respectively) and a distal end cross section of dimensions 90×65 mm, slightly flattened on the ventral side as observed in the biological trunk. A loft with quadratic tapering (a parabolic profile with its minimum at the trunk's tip) connects the two ends, both consisting of two 8 mm thick plates built monolithically with the lattice (see Fig. S1 and S2). The so-obtained outer shape, which defines ELEANOR's body, will be hereafter referred to as the *bounding trunk*.

### *Properties of the lattice-based trunk*

Within the bounding trunk, a lattice is generated with a triply periodic minimal surface. The unit cell size and the strut diameter of the lattice (measured at the midpoint) are 25 mm and 2 mm, respectively (detailed in Fig. 1). The effective volume of the tessellated structure is about 11% of the volume of the bounding trunk (estimated as 0.0013 m$^3$ considering also the two end plates, and 0.0116 m$^3$, respectively). Since only one material is used, mass and volume rescale accordingly: a full, non-tessellated trunk would have a mass of 12 kg, whilst ELEANOR's total mass is 1.36 kg (including the end plates, measured with a laboratory scale).

### *Fabrication: 3D printing and seamless connection of large-scale lattice-based parts*

The material used to build the robot is the resin Elastic DL160A® (Photocentric Ltd, UK), purposely developed for this work to enable large scale 3D printing by the LC Magna (Photocentric Ltd, UK) having a build volume of 510×280×350 mm. The robot's body is printed in one run in three parts, with the proximal and the middle segments having equal length (see Fig. S3). This aspect is essential for printing success due to the behavior of uniform lattices when subjected to the delamination forces from the vat during the printing process. The magnitude of such peeling forces depends on parameters such as resin viscosity, flowability, and build plate retraction speed, but more crucially on the surface area of each printing layer. For an elastomeric resin this poses a unique issue in which the stretchability of the bulk model scales positively with the total number of unit cells present in the Z-axis. As the model grows, the distance the platform must retract – before the layer successfully delaminates – must also grow. Indeed, during the preparation of the final fabrication process it was found that a Z-axis lifting distance of 24 mm between each layer is necessary to enable successful fabrication.

The Elastic DL160A® mechanical properties are a second crucial factor. To build the trunk, a soft and highly deformable material is required, yet strong enough to withstand both the forces of tendon-driven movement and the printing process itself. The development of the material was focused on maximizing its tensile strength and elastic modulus in the 'green' state, through selection of its constituent monomer/oligomer starting materials. It was also found that the resin had a cure depth of up to 400 μm layers on LC Magna without noticeable light-bleed in the XY-plane, which drastically reduced the total number of layers and therefore the total print time from approximately 24 to 14 hours.

After post-processing, the robot is finally obtained by gluing the three parts at their interfaces, precisely at the lattice nodes located at their ends. In the preparation of the files for the printing, the tessellated geometry was sectioned purposely splitting such nodes into two identical halves, providing a suitable surface for gluing. Hence, the connection was possible without introducing discontinuities in the body, achieving the desired continuity.

***Biomimetic tendon-driven actuation in lattice architecture***

The thousands of muscle fascicles in the elephant trunk are organized into a few muscular groups: longitudinal, transversal, radial, deep oblique and superficial oblique muscles (*32–34*). In ELEANOR, tendon-driven actuation mimics a subset of such groups which represent more than 50% of the total volume of an elephant trunk (see (*4*)): deep and superficial oblique muscles (DO and SO, respectively), and longitudinal muscles (L). Oblique tendons are placed in the ventral region of the robot, while longitudinal tendons run dorsally. In total, the robot is actuated via six tendons that run along the entire length of the trunk, placed symmetrically with respect to the sagittal plane, as depicted in Fig. 2A. In the following, the subscripts L and R will denote whether the tendon path begins on the left or the right side of the base of the trunk, respectively.

The deployment of the tendons does not require dedicated channels or precise positioning procedures to route them. Tendons consist of Kevlar fibers enclosed in silicone tubes to protect the lattice from wear. The friction between silicone tubes and lattice has little impact on the maneuvers because of the inherent damping of the robot structure, which absorbs the energy released during stick-slip events. Each tendon is pulled by a motor placed on a fixed frame at the base of the robot and is anchored at the distal plate of the robot.

As expected, we observe that the initial tension imposed on the tendons affects the overall mechanical response upon a given actuation input. For the experimental assessment, after setting the robot at starting pose, the trunk is pre-compressed by ~7% of its length. This allows for avoiding slack tendons and pursuing symmetrical behavior between left and right. Nonetheless, perfect symmetry is hard to obtain, as highlighted later.

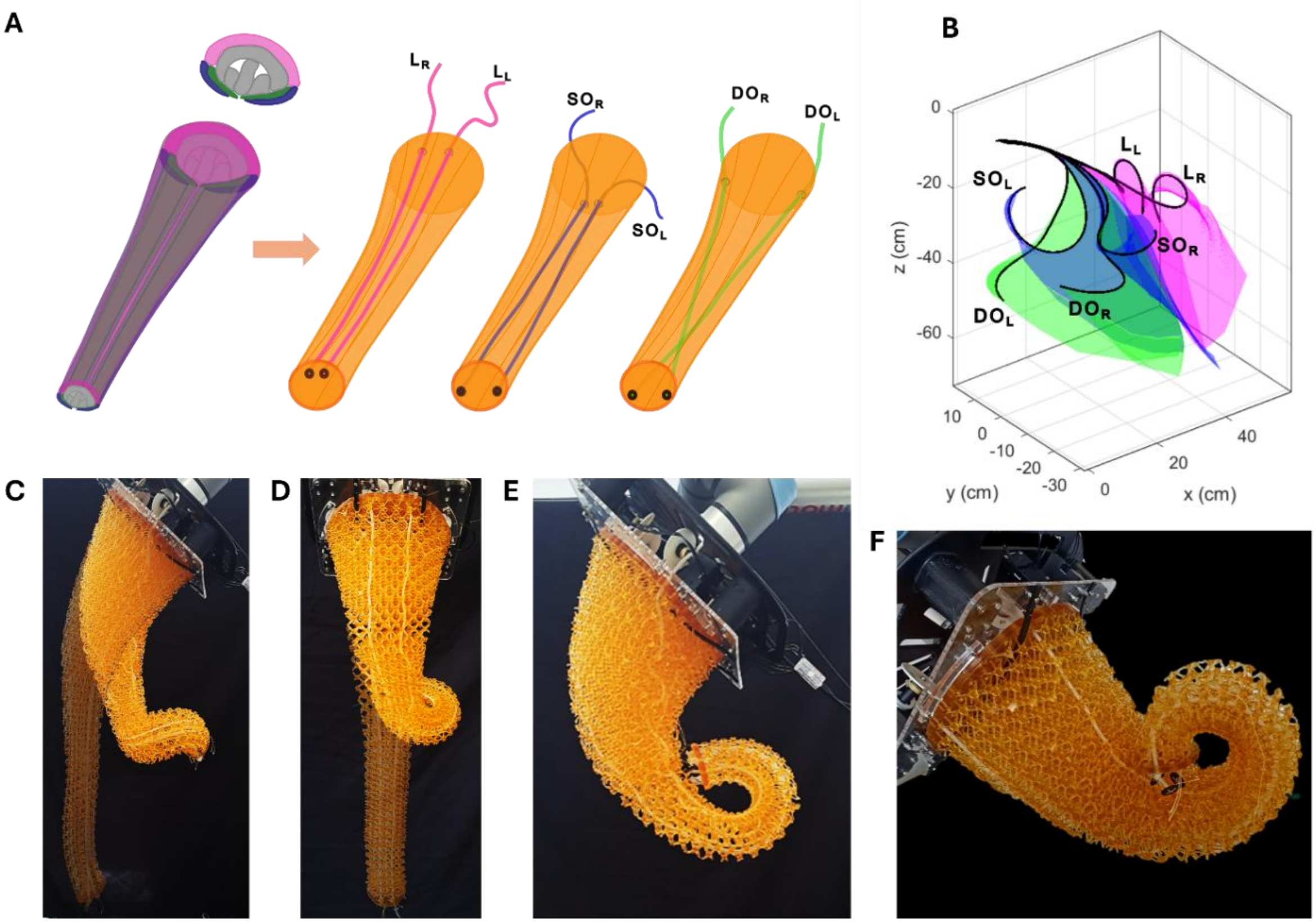

**Fig. 2. Deformability upon tendon actuation.** (**A**) Pictorial representation of the muscular groups considered for ELEANOR's actuation: longitudinal (pink), superficial oblique (blue) and deep oblique (green). (**B**) Deformed configurations achieved upon pulling tendons individually. (**C**) Side view and (**D**) front view of ELEANOR performing non-planar bending upon use of $SO_L$. (**E**) Planar bending downward is achieved by symmetric actuation of superficial oblique tendons. (**F**) Planar bending upward, achieved by symmetric actuation of longitudinal tendons.

### *Homogenized properties: ELEANOR vs. biological trunk*

Although tuning the properties of architected structures and, more in general, of metamaterials is a current promising trend in soft robotics (*43*), in this work we adopt a different approach. Preprogrammed deformations along the axis of the robot (*i.e.*, by implementing stiffness gradients in the lattice) are not pursued since the biological trunk does not work according to that principle. Although its structure is non-homogeneous, its fundamental components (muscles, connective tissue, skin) are consistently present along its entire length (*4*, *32*). Any portion of the trunk (excluding the fingers) possesses the hardware to contract, bend, twist: there is no specialization for a specific deformation mode along the trunk.

In this work, we focus on capturing and mimicking main characteristics of the elephant trunk *(i.e.*, proportions and dynamic properties), rather than the detailed anatomy, to achieve a bioinspired continuum arm at the large scale. Hence, let us consider the elephant trunk as a homogeneous structure, made of a single material (with the properties of muscular tissue). It can be roughly described as a soft, long beam constrained to a moving frame (representing the head) and which, when not actuated, is free to naturally swing. During previous experimental activities with elephants (*41*) we observed (yet not quantified and reported) that elephants, if not involved in any task, tend to let their trunk swing freely at relatively low frequency. Since we cannot provide a quantitative experimental assessment of this oscillation at present, we recur to modeling for an estimation of such frequency. For comparison, but also for the purpose of validation, we use the same model to simulate the oscillations of ELEANOR, which are also experimentally studied. The model used is two-dimensional, undamped, and is based on the discretization of a generic trunk as a series of beam elements. For the purpose of this work, both the biological and the artificial trunk are modeled through homogenization, also because our knowledge of the biological properties is limited to the apparent elastic modulus and density (see Introduction). Reducing the lattice to a homogeneous material, the apparent density is ~90 kg/m$^3$. The apparent elastic modulus, estimated through simulation, is approximately 6 kPa under compression and 25 kPa under elongation (see Materials and Methods).

Movie S1 shows ELEANOR swinging freely after being released upon an imposed initial displacement and its behavior as computed by the model (see also Fig. S7). Tracking the distal plate during ELEANOR's oscillation revealed a period of ~0.9 s, which corresponds to a frequency of 1.1 Hz. The value found through modeling is ~1.2 Hz, proving that the model is sufficiently accurate for our purpose. Then, it is legitimate to use the same model for the biological trunk, assuming for it an elastic modulus E = 938 kPa (*39*) and a density ρ = 1180 kg/m$^3$ (*40*) (as summarized in Table S1), obtaining a frequency of ~0.9 Hz. Hence, ELEANOR has a dynamic behavior plausibly compatible with that of the biological trunk. This analysis is based on results of a simplified model relying on data extracted from existing studies. A more accurate comparison would require a deeper knowledge of the mechanics of the trunk and of its behavior in relaxed and non-relaxed state. However, this kind of data is hardly available due to the objective difficulties in performing tests with the animals. Herein our goal is to roughly match the biological system's properties. Let us remark that ELEANOR is smaller than its biological counterpart. If we built a real size trunk with the same lattice used for ELEANOR, that system would obviously have different frequencies. Focusing merely on the apparent material properties, it is useful to consider the ratio $c = \sqrt{E/\rho}$, which is the velocity at which a wave propagates through the material (in the one-dimensional case). Such quantity was considered in another study related to the elephant trunk (*40*) to speculate about its behavior. Herein it is used for a rough comparison between the artificial and the biological. In this case, we obtain ~14 m/s and ~28 m/s for artificial and biological material, respectively. This means that, to preserve eigenfrequencies, a real-scale robotic trunk (that is, having the same geometrical parameters of an elephant trunk) should be built with stiffer and/or lighter lattice than the one used for ELEANOR.

We address this matter because the dynamic characteristics of continuum arms, especially if soft, have not been adequately discussed yet, while this is one of the factors that hinder effective deployment of this class of systems. Since they are usually actuated at relatively low speed, they often suffer from undesired vibrations. Quantitative reports of such vibratory behavior are scarce in the literature. However, long, slender continuum robots require time intervals in the order of minutes to perform a task such as pick-and-place (see, *e.g.*, (*27*)) or vibrate visibly upon actuation (*e.g.*, (*44*)). Damping must be considered, however, to the best of our knowledge, no information is available about damping of the elephant trunk's tissues, and in this work the design is not based on a prescribed damping factor. We only note that ELEANOR's structure is underdamped: Movie S1 shows that ELEANOR performs a few complete oscillations before reaching static equilibrium. A positive outcome of its low eigenfrequencies but high damping is that its behavior is only marginally affected by impulsive loads. For example, even when elastic energy is accumulated locally due to the friction between the tendons and the lattice, and suddenly released due to slipping between the two, the robot does not respond with fast oscillations. Peak forces are reached slowly, and operations can proceed smoothly. To better show the true dynamic behavior of the robot across all experiments, all the supplementary videos are provided at real speed.

***Large deformations enabled by structural compliance***

A Fiber Bragg Gratings sensor (FBGS) is used to perform shape reconstruction: the thin sensorized fiber (Ø 0.3 mm) is encapsulated in a PTFE tube and deployed along the centerline of the robot. During the movements, its shape is recorded and subsequently analyzed. The objective is to quantify the amplitude of the trunk displacements, and to explore the deformed shapes exhibited upon both individual and combined action of the tendons. First, we show the deformations achieved when tendons are used individually: Fig. 2B shows such deformations for each of the six tendons, to which the greatest contribution is clearly given by the distal portion of ELEANOR. Figures 2C and D show, as an example, the deformation obtained through action of SOL, overlapped with the starting pose: the proximal portion of the trunk remains almost undeformed while the distal plate undergoes a vertical displacement of 300-600 mm depending on the actuation input. Based on FBGS data the smallest volume within which ELEANOR performs all the movements has dimensions equal to 0.58×0.48×0.75 m, as shown in Fig. 2B. Clearly, all six configurations imply out-of-plane deformation, as no tendon lies on the sagittal plane of the robot. Planar bending is achieved by simultaneous action of a pair of tendons (left and right), as shown in Fig. 2E and F, in which ventral and dorsal bending results, respectively, from using the two superficial oblique and the two longitudinal tendons. In these cases, self-contact occurs, as the distal plate reaches the lattice, creating a closed loop.

It can be noted that configurations achieved on one side of the trunk do not exactly correspond to those on the opposite side. Sagittal symmetry is compromised by multiple factors simultaneously. Besides the potential asymmetries in the initial pretension mentioned above, other factors are difficult to mitigate: the lattice, although symmetric by design, is affected by local inaccuracies and deviations from its nominal dimensions stemming from the fabrication process (*45*) (see Supplementary Materials). Moreover, the behavior is also influenced by the friction between the tendons and the lattice, which is configuration dependent. Although these effects are not specifically quantified in the present study, given the observed uniform distribution of such inaccuracies ELEANOR can be considered as a homogeneous structure. Another aspect worth mentioning is related to the material dissipative effects. Figures 3A, D and F show FBGS trajectories during a full actuation cycle: the robot starts from its home pose, reaches gradually its final deformed state, and after 10 s tendons are released. As shown in the figures, the home pose is not perfectly recovered due to residual strains throughout the lattice. During the experimental analysis we found that the distance between the center of the distal plate at the home pose and at the final pose ranges from ~20 mm to ~100 mm (that is, an error ranging between 2% and 11% of the trunk length) depending on the provided actuation pattern, and also on other factors such as room temperature. While we suggest that this could be mitigated by tuning the material properties, or compensated through antagonistic actuation inputs, we also remark that residual deformations can be easily corrected by a human operator interacting with ELEANOR. Movie S2 shows that such interaction is safe and fully compliant, owing to robot's mechanical properties.

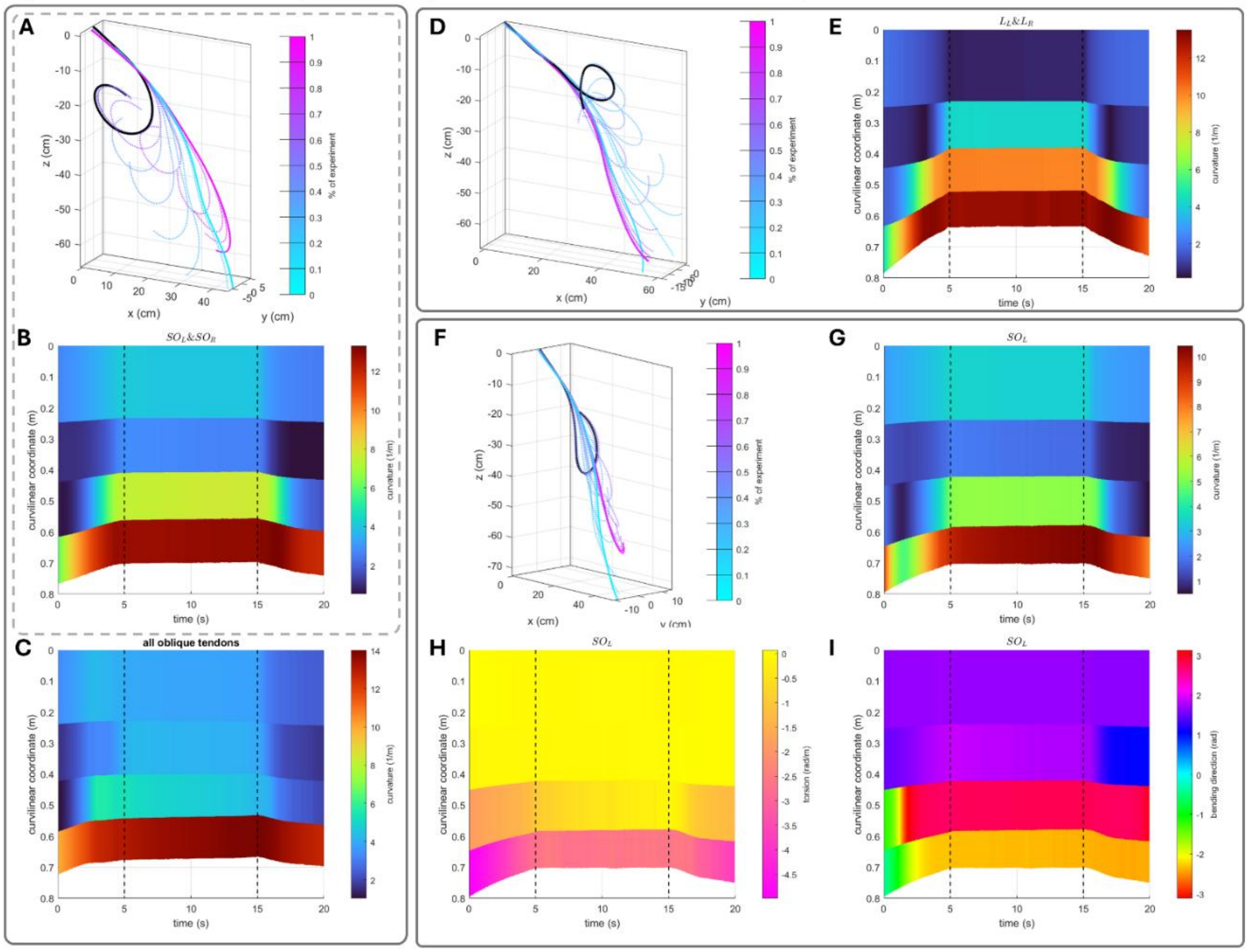

**Fig. 3. Mechanical behavior throughout complete actuation cycles.** (**A**) Evolution of the configuration of the robot during planar bending downward, achieved by simultaneous use of $SO_R$ and $SO_L$. Data are obtained by means of FBGS. (**B**) Curvature and contraction of the trunk upon actuation of $SO_R$ and $SO_L$, the former being denoted by the colormap, the latter being represented by the shortening of the segment. Data are obtained via motion capture. (**C**) A similar trend is observed upon actuation of all the oblique tendons, with greater values of contraction. (**D**) Evolution of the configuration during planar bending upward, achieved by simultaneous use of $L_L$ and $L_R$. (**E**) Curvature and contraction exhibited upon such actuation. (**F**) Evolution of the configuration upon actuation of $SO_L$. The corresponding curvature, torsion, and bending direction, are shown in (**G**), (**H**) and (**I**) respectively. Compared to the case of symmetric actuation (i.e., $SO_R$ and $SO_L$), contraction occurs with the same values, whilst smaller curvatures are achieved.

During the experiments, the FBGS is left free to slide along the robot body to prevent damaging it due to possible kinking. This impaired quantification of ELEANOR contraction caused by the tendons. To investigate this aspect a motion capture system is utilized. The trajectories of five equally spaced rows of markers placed along the robot body (see Materials and Methods) are recorded during robot operation. Data are postprocessed and visualized to highlight the longitudinal compression and curvature (for comparison vs. FBGS curvature data) of the four considered segments. As expected, results show that the amount of contraction increases from the proximal base to the distal end of the robot. Figure 3B shows that the robot contracts in total by ~9% of its home-pose length upon symmetrical actuation of the superficial oblique tendons, while the proximal, mid-proximal, and mid-distal segments contract by ~5%, ~10% and ~15%, of the initial length upon actuation, respectively. The plot shows that contraction of the distal part appears smaller than that of the previous segment (being ~7.2%) but one must account for the initial greater contraction that occurs to this segment owing to the initial compression imposed. In fact, at the home pose the robot points downward, and the pretension applied to the tendons determines the distance of the distal plate from the ground. The entire body is subject to gravity, which causes a larger initial compression at the distal region. If the length of the distal segment is considered in the rest configuration, such segment undergoes ~26% contraction. The same considerations hold for other actuation inputs (see Fig. 3C, E, G). What was found for the contraction is consistent with the curvature distribution: while the proximal region exhibits a relatively small curvature (4 $m^{-1}$ at fully actuated state), the distal region is highly curved, reaching curvatures up to 10-13 $m^{-1}$, and with an increasing trend from proximal to distal section.

Concerning out-of-plane shapes in ELEANOR, they do not stem from torsion alone, but they also arise whenever the bending plane rotates along the trunk – that is, when the bending plane is not uniform from base-to-tip. In free movement, this rotation is typically distributed: Figure 3H shows how an oblique tendon ($SO_L$) keeps a modest torsion (-0.57 ± 0.98 rad) but introduces a gradual base-to-tip reorientation ([1.85, 2.01, 3.09, -2.41] rad, with counterclockwise direction, see Fig. 3I) that yields a gentle 3D drift even.

### *Whole-body grasping*

Achieving large displacements and curvatures *per se* does not enable whole-body grasping. ELEANOR's design allows for curvatures that are not only large, but also with a suitable distribution for adapting to an object contour and wrapping it. To investigate such capability, the grasping was tested with various objects, some of which purposely 3D printed with assigned dimensions, and others selected from our daily working environment. Herein the aim is not to target high payloads and the objects' weight is in the range of 0.1 – 0.2 kg. Results obtained from trials with cubic and cylindrical objects with dimensions ranging from 75 mm to 150 mm (detailed in Table S2 and selected based on the experiments performed with animals (*32*)) are shown in Fig. 4. Given an object and an actuation input, involving one or more tendons, the grasping is tested 10 times, and classified as successful if *(i)* the robot wraps the object firmly, and *(ii)* the arm holds the object until after the recruited tendons stop pulling. The experimental protocol is provided in Materials and Methods.

As an example, Fig. 4A and B show ELEANOR grasping a cylinder (Cyl-S, d = 75 mm) and a cube (Cube-S, L = 75 mm) upon actuation of $SO_R$. The lattice establishes a contact which depends on the shape of the object. In the first case, the contact with the cylindrical object occurs on a set of lattice nodes. The contact loci can ultimately form a continuous line if the pulled length is such that the lattice nodes contact each other upon compression. Differently, when grasping a cube, the lattice conforms to its surface despite the edges. Figure 4C summarizes the success rate for these two objects: results are shown for each tendon, left and right and for the case of symmetric actuation. As previously mentioned, performance is affected by the initial tension in the tendons, which explains the differences between left and right. However, the most successful grasping modes imply planar bending, that is, symmetrically actuated tendons. With such actuation input, ELEANOR can grasp larger objects: Fig. 4D shows dorsal grasping of a cylinder (Cyl-M, d = 100 mm). Due to the size of the object, the trunk wraps around with a smaller contact area than in the case of smaller objects. Indeed, the size of the objects dictates the curvature of the artificial trunk, given its deformability. For example, Fig. 4E shows the results obtained by means of FBGS for the case of ventral grasping of Cyl-S, where the distal region, which wraps the object, achieves a curvature radius of ~70 mm, which corresponds to the sum of the radius of the object and the distance between the FBGS and the ventral surface of the robot. More results, with different objects and for different actuation inputs, are reported in Fig. S12. Overall, it is observed that ELEANOR grasps objects whose size is in the range ~70 –100 mm. Highlights of the grasping capabilities of ELEANOR are shown in Movie S3.

While performing grasping with out-of-plane deformations, the effect of the added distal load and contacts shifts the rotation of the bending plane toward medial/distal regions, making the out of plane component more pronounced and often accompanied by higher torsion (x4.31 with respect to no-object case) – see Fig. S13. These observations indicate that an out-of-plane configuration is best interpreted as the combined effect of two routes – (i) a spatially varying bending direction and (ii) material twist – whose relative weight is set primarily by actuation pattern (paired vs. single, longitudinal vs. oblique) and loading condition (free vs. grasping), while small pretension, friction, and fabrication tolerances modulate the observed 3D departures from planarity.

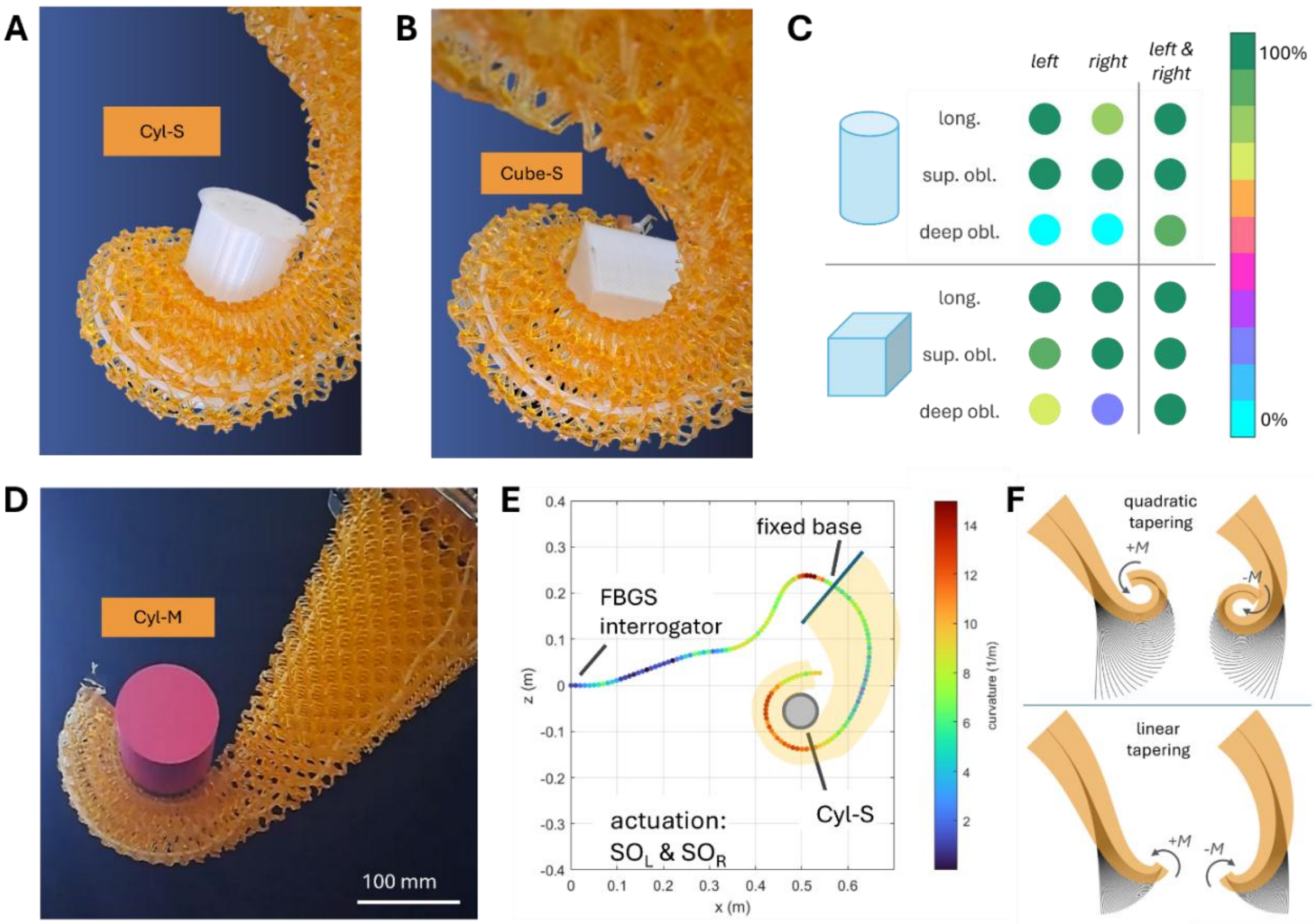

**Fig. 4. Assessment of ELEANOR's grasping capabilities with regular objects.** Upon action of $SO_R$, grasping of (**A**) cylindrical object (Cyl-S, d = 75 mm, h = 150 mm), and (**B**) cubic object (Cube-S, L = 75 mm), with (**C**) success rate out of 10 trials. (**D**) ELEANOR grasping through dorsal planar bending a larger cylinder (d = 100 mm, h = 240 mm). (**E**) Distribution of curvature as obtained by means of FBGS when grasping Cyl-S with ventral grasping. (**F**) Advantage of quadratic tapering over linear tapering in terms of effective curvature distribution for grasping.

ELEANOR can be used as a tool to better understand the biological trunk, and an important point can be raised relating to grasping. In a recent work, researchers investigated how the various muscular groups contribute to certain motions of the elephant trunk (*46*) and concluded that, for the trunk to perform successful ventral grasping, radial muscles play a fundamental role: their action compensates the longitudinal contraction caused by the ventral muscles, enabling grasping otherwise unfeasible due to incomplete wrapping around the object. While we do not question the validity of that study, we argue that successful whole-body grasping does not strictly require the implementation of radial muscles in a robotic trunk, as proved by ELEANOR. What looks like an inconsistency between our work and the results in (*46*) is actually not related to the number and the type of muscular groups in the system: the culprit here is solely the geometry of the bounding trunk. Indeed, it is a commonly adopted assumption that the elephant trunk has the shape of a truncated cone as in (*46*) (but also in (*36*), (*40*), (*37*), (*14*), (*47*), and (*48*)). Hence, we suggest that assumptions on the geometry of the biological trunk should be formulated cautiously. However, from the engineering viewpoint, an artificial trunk with linear tapering would be disadvantageous. The quadratic tapering of ELEANOR allows for the involvement of its entire distal region in the grasping owing to a convenient curvature distribution. This can be easily demonstrated also through modeling, comparing the planar bending of two tapered beam-like structures with linear and quadratic tapering, subject to the same constraint and loading conditions: Fig. 4F shows the difference between a trunk with quadratically decreasing radius and a conical trunk upon the same boundary conditions (see Supplementary Materials). Also, in relation to grasping, it is possible to draw similarities between ELEANOR and its biological counterpart. ELEANOR performs bending with a curvature distribution that is similar to that exhibited by the considered African elephant trunk. In the case of ventral grasping achieved by simultaneous action of the two superficial oblique tendons, shown in Fig. 4E, the distal part wraps around the object. The curvature is relatively low from the fixed base up to ~60% of the trunk length (i.e., in the proximal and middle regions), with a steep increase in the remaining length. The same trend was observed during a test with a living African elephant grasping a cylindrical object (*cf.* Fig. 3B in (*32*)). This similarity arises despite the unavoidable contraction of ELEANOR's body, which does not impair grasping. This can be explained not considering the total contraction of the trunk (*i.e.*, from a relaxed length of 85 cm to a contracted length of ~70 cm), but the distribution of such contraction along the trunk: since the lattice has the same stiffness throughout the bounding trunk, each section contracts based only on its number of unit cells (*i.e.* on the size of the cross section). Once again, global behavior is determined by the tapering law.

Another worth mentioning aspect is that a body made of lattice facilitates the interaction with irregular objects, owing to local deformation of the unit cells in contact with the objects. ELEANOR's versatile behavior is shown in Fig. 5 by depicting grasping tasks of different kinds of objects having irregular shapes.

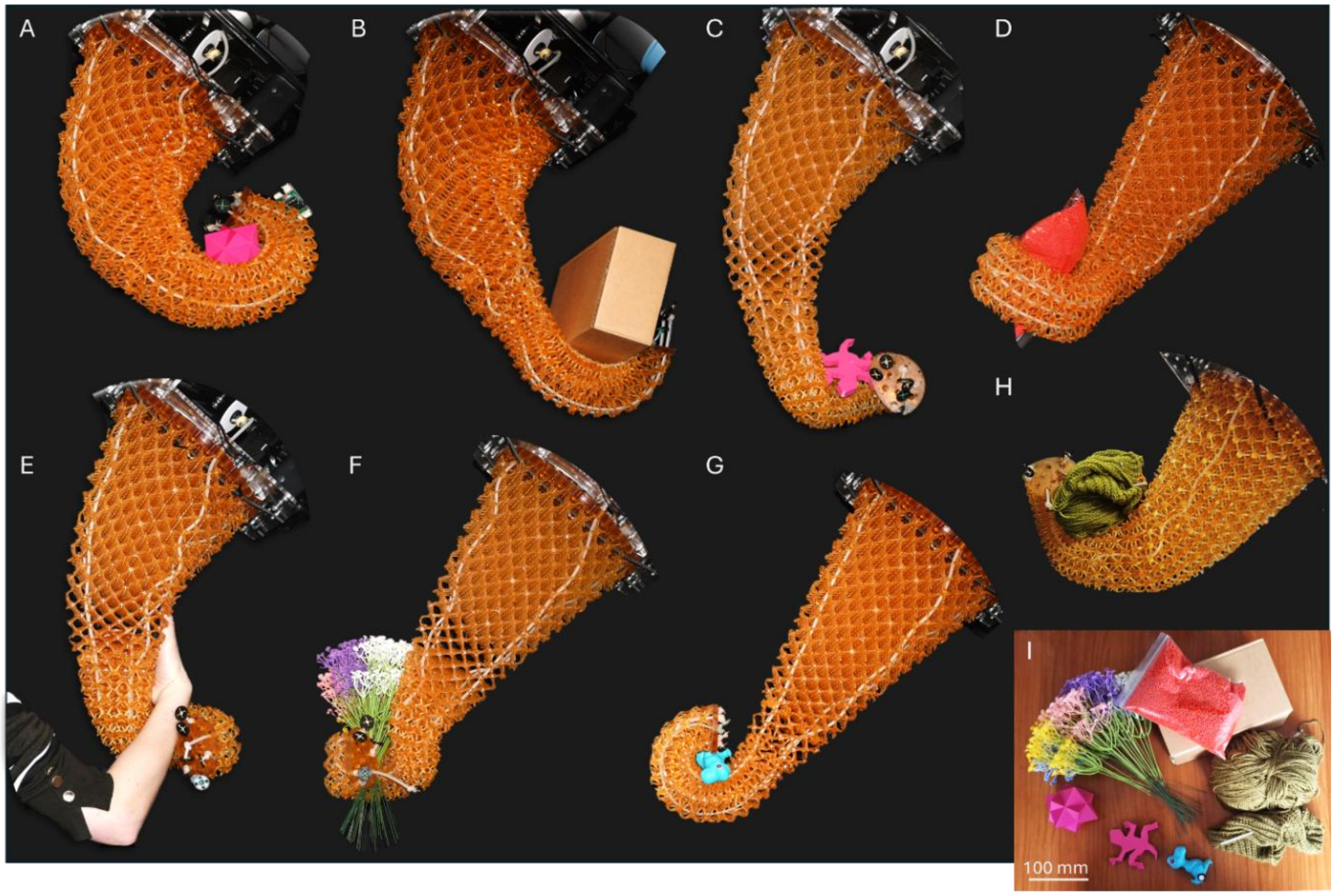

**Fig. 5. Examples of grasping with non-standardized objects.** ELEANOR interacting with: (**A**) a star-shaped object, (**B**) a box (210×130×80 mm), (**C**) a flat irregular object shaped like a lizard (thickness 12 mm), (**D**) a bag containing granular material, (**E**) a human that pulls ELEANOR during its operation, (**F**) a flower bouquet, (**G**) a toy, (**H**) a crochet work. All the objects are shown in (**I**).

***Durability***

An aspect that deserves attention is durability, especially if the artifact is monolithic, meaning that part replacement is not feasible. This holds true not only to ensure correct behavior of the robot in operational conditions, but also in view of developing future environmentally sustainable machines. ELEANOR maintained its mechanical integrity after more than ~1000 tests of various kinds (*i.e.*, grasping experiments, contact with a human being, large bending and twisting, etc.), carried out over one year. Then, a few lattice unit cells failed (mostly at the transition between the strut and the node of the lattice) but were easily repaired either with commercial glue or using the curable resin that constitutes its body. Moreover, despite the failure of a few struts the trunk continued to perform as expected.

## Discussion

Results demonstrate that combining a bioinspired approach with an architected design at large scale is a valid path for building a continuum arm capable of performing trunk-like behaviors, which are safe for human-robot interaction. Achieving uniform, and repeatable large lattices by 3D printing enables to harness structural, in addition to material-based, compliance. The smooth coupling of the tapered volumetric structure with tendon-driven actuation (mimicking longitudinal and oblique natural muscle groups) yields a robust yet simple system that shows main traits of the natural organ, like its dynamic response and curvature distribution. Indeed, creating a continuous interface with external objects (by compression of the lattice body) has a key role for adaptability and versatility of prehensile tasks. Whole-body grasping is demonstrated with objects having size up to 12% of the robot's length (85 cm), with ELEANOR being capable of large curvature (up to 14 m-1). The distal end achieves three-dimensional displacements equal to 70% of the robot's body length. Such large deformations, adaptation to diverse object geometries and hardness, and durability, support the proposed strategy as a viable pathway for durable and adaptable soft systems at the large scale.

***The path ahead***

As shown, ELEANOR can grasp objects effectively without relying on elongation (although it has been emphasized as a crucial capability for grasping in the elephant (*46*)). Indeed, the possibility to elongate would offer a longer portion of our robot body to wrap objects, and also to reach them (*32*). Certainly, this would come at the cost of increased complexity in the actuation and in the control. While longitudinal and oblique tendons can be deployed straightforwardly along the length of the robot, placement and anchoring of radial muscles would be of much more difficult implementation. Moreover, they would require motors that should be hosted within the body structure.

A further remark concerns the hydrostat nature of the biological trunk – a feature that is not implemented in this work. ELEANOR's body contracts longitudinally while not expanding radially. Such expansion, if present, would even play against the grasping performance: an increase of the radius of the cross section would correspond to a decrease of available space in the concavity of the curved portion for hosting the external object. Therefore, implementing muscular hydrostat behavior might not be a must-have in this kind of architected robot. However, a deeper investigation in this direction is needed to better understand whether such a characteristic could bring advantages for robotic manipulation.

It must be mentioned that the morphological model that is used in this study derives from a single individual (*32*, *42*), and the behavioral grasping benchmark (*32*) is from an elephant with a previously severed trunk tip. This represents a limit for generalization. However, experiments involving elephants are difficult to perform (e.g. availability of both animals and parks that guarantee animal welfare and rigorous experimental protocols), and for anatomical studies organs from naturally deceased elephants are needed.

Concerning the purely technical aspects, there are open challenges in the fabrication of monolithic soft systems at large scale. In this work, a 3D printer with a large build volume is used and ELEANOR is printed in three parts concurrently. Fabricating one artificial trunk monolithically is possible using a larger printer, in principle. However, the increase in unit cells, therefore total possible deformation, means that a one-piece build quickly becomes uneconomical and prone to failures due to long Z-axis retract distances. This problem exists for most tall lattice structures made from soft, elastic material. However, 3D printing is a widely investigated field, and it is reasonable to expect new outcomes in the short term, depending on the demand. Meanwhile, it is useful to account for any inaccuracy introduced by the fabrication process, which might unpredictably alter the characteristics of the designed artifact. This aspect is still an open issue in soft material printing (defects might invalidate any structural optimization, however sophisticated) and it motivates an approach based on the global properties of the biological trunk, making local defects irrelevant as they influence global behavior in a negligible way.

Finally, while this study was focused on the large-scale design and fabrication, it will be crucial to integrate extero/proprioceptive sensing capabilities throughout the whole trunk to investigate innovative closed-loop control methodologies as well as integrating the tip of the trunk for pinching tasks. Towards this objective we are developing monolithic techniques to achieve a sole construct for body-actuation-sensing, like recently shown in a soft monolithic gripper (*49*) inspired from the distal portion of the natural trunk (with the opposing fingers at the tip), and that can reach objects by elongation. The open challenge is to evolve such approach to achieve whole body grasping in addition to pinching tasks, without dramatically increasing structural and robotic complexity.

***Performance-driven design, or biomimetics first?***
In summary, instead of chasing prescribed deformations and performance, we aimed at capturing the trunk's fundamental features and we evaluated the outcome *a posteriori*. We traded structural optimizations, stiffness gradients, preprogrammed mechanical responses, with a simple design approach, and eventually verified that some of the observed elephant-like behaviors could be achieved with our meter scale system. Optimization of local properties can be postponed to a further, subordinate design stage, accounting for the limits due to the fabrication process as well.

We conclude by suggesting that to address effective, versatile, large-scale manipulators, taking inspiration from the elephant trunk, optimizing only a specific stiffness should be avoided. This is particularly important when considering embodied intelligence (*50*) as key in the quest for future effective robots, thus where the body mass has a role. If we think in terms of dynamic modeling, we realize that inertial forces appear in the equation with due reason, especially for large dynamic systems such as elephants. We should embrace the idea that the elephant trunk might work amazingly not despite, but because of all its properties, size and weight included. Perhaps, standing over the engineer's conventional approaches, and unconcerned of expectations on robot performance, there is a reason also for those size and weight, that is worth chasing.

## Materials and Methods

***Robot design: dimensions of the bounding trunk***
From the fabrication standpoint, reproducing the elephant trunk at 1:1 scale would have been feasible, but it would have posed challenges related to the needed facilities for the experimental evaluation of the performance. An African elephant's trunk, in fact, is ~1.4–1.8 m long (*32*, *40*). For our first large-scale robotic trunk, it was preferable to work on a biomimetic robot smaller than its biological counterpart. While the length was assigned based on the build volume of the available 3D printer, other dimensions stemmed from the discrete nature of the lattice, which imposes constraints on the deployment of the tendons. Being the lattice unit cell size 25 mm, in fact, it was convenient to ensure that at least two complete unit cells were present along the height of the cross section, and three along the width, to allow for biomimetic tendon routing. The geometry shown in the Supplementary Materials (Fig. S1) allows for satisfying these conditions.

***Lattice: topology, dimensions, stiffness***
Lattice is generated by means of the software 4D_Additive (CT Core Technologie GmbH, DE) within the volume of the bounding trunk. A built-in feature of the software is used to create a I-WP triply periodic minimal surface resulting in the lattice visible in Fig. 1. Due to the build volume of the printer, the trunk is sliced into three parts. The slicing is performed at the midpoints of nodes lying on the same section, in order to facilitate gluing of adjacent portions.

To investigate the mechanical response of the printed lattice, cubic samples featuring the same I-WP-based tessellation as ELEANOR are simulated under both compression and elongation in SOFA (*51*) with a tetrahedral mesh applying the imposed displacement incrementally (1 mm/min for a simulated cube with 50 mm edge). The obtained stress–strain curves are averaged for the computation of the apparent elastic modulus. For validation purposes, compression tests on 3D printed cubic samples with the same dimensions are performed (Fig. S5). Under compression, the apparent elastic modulus is 6 kPa, while tensile simulations yielded a value of 25 kPa, indicating a stiffer response in elongation (Fig. S6).

***Dynamic modeling of trunk-like structures***

To estimate the oscillatory behavior of trunk-like structures, we implement a simple numerical model which fits the purpose of this work. An accurate calculation of the nonlinear eigenmodes of the considered trunks is out of scope. Here, the objective is to assess whether ELEANOR and the biological trunk share similarities in terms of dynamic behavior. Hence, for this work we use a model based on the following assumptions and simplifications: *(i)* it describes only oscillations occurring on the sagittal plane of the trunk, that is, it is two-dimensional; *(ii)* the cross-section of the trunk is considered circular; *(iii)* damping is not modeled; *(iv)* linear elastic material is considered.

The generic trunk is modeled as a series of two-node beam elements, with each element described by its corresponding stiffness and mass matrices. The first node of the first element is fully constrained, and the only external load applied is the gravitational force, simulated as a distributed load acting along the vertical direction. No other external loads are applied. The trunk is inclined by 10 degrees with respect to the vertical axis and left free to swing upon gravity for 2 s.

The problem is implemented in MATLAB R2023b (The MathWorks Inc, US). The equations of motion are solved by means of the solver *ode45*. At each step, the stiffness and the mass matrices of the elements are updated based on the result obtained at the previous step, *i.e.*, the model accounts for large displacements.

This model is used to simulate the behavior of two trunks having the approximate dimensions of a biological trunk of an adult elephant and of ELEANOR, for comparison. To ensure that the model is sufficiently accurate for our scope, we compare the modeled and the experimental oscillation of ELEANOR: we track the trajectory of the distal plate using an open source software (Tracker Online (*52*)). We find that modeling results are in acceptable agreement with experimental evidence (see Fig. S7 and Movie S1).

The same stiffness matrix generated for this problem can be used to perform quasi-static simulations. We used it to test different combinations of geometric and material properties, obtaining results such as those shown in Fig. 4F.

We report a detailed description of the model in the Supplementary Text.

***3D printing: material development and assembly of the 3D printed parts***

The Elastic DL160A® resin was purposely developed for this work within the PROBOSCIS project (*53*) and subsequently commercialized (see datasheet (*54*)). Through several printing iterations, we identified the optimal printing parameters and the best layout to achieve high-quality printed lattice while facilitating the final assembly of the robot body.

Assembly of the three printed sections required a strong adhesive at the two interfaces created when slicing the original monolithic model into three parts. Several adhesives were proposed which successfully bonded the sections, however introduced two planes at which the mechanical properties differed from the bulk, creating a mechanical discontinuity. To not affect the overall deformation characteristics, using Elastic DL160A® as a UV adhesive is the best option. Resin is applied to each lattice node at the interfaces using a syringe, and a UV torch (405 nm) is used to cure the parts together. This method ensures an elastic mechanical bond, but also a chemical bond as the resin reacts with uncured terminal chains on the surface of the material. A second UV post-cure step is then used to ensure full bonding between the sections of the completed robot. This results in a final part that demonstrates continuity and mechanical integrity.

***Experimental setup and protocol***

ELEANOR's proximal base is constrained to a rigid platform that hosts six motors to pull and release the tendons. The used motors are Dynamixel XM430-W210, (ROBOTIS Co., Ltd, South Korea). The platform is fixed as the end-effector of a rigid-linked robot UR5e (Universal Robots A/S, Denmark) that holds the robot and sets its orientation with respect to the horizontal plane. The platform (shown in Fig. S9) is built through 3D printing.

Shape reconstruction was performed by means of a Fiber-Bragg Grating Sensor (FBGS Technologies GmbH, Germany). The hardware of the system consists of a sensorized optical fiber (1.5 m long) and of an interrogator box to which the fiber is connected. Signals from the interrogator are elaborated by means of ILLumiSense® v3.1.x software, streamed to Shape Sensing® v1.5 software, and saved for post-processing in MATLAB R2023b (The MathWorks Inc, US).Due to its hardness and its small diameter (0.3 mm), the sensorized fiber is deployed along the centerline of the robot encapsulated in a thin PTFE tube (diameter smaller than 0.8 mm). During all the experiments, the fiber is allowed to freely slide along the robot, to prevent damaging it due to possible kinking. Therefore, the distal end of the fiber protrudes out of the top plate of the trunk, with the protruded length depending on the configuration of the robot. The other end point of the fiber is connected to the interrogator, placed on the same table that hosts the UR5e. The portion of fiber between the interrogator and the platform is fixed to the table and to the platform, and it is identified during post-processing as the length between the fixed point (the connector) and the point with the greatest z-coordinate (that is, the highest point) corresponding to the entrance point in the trunk. This is necessary to reduce errors as much as possible in shape reconstruction. Despite correct usage of the FBGS, the coordinates of the reconstructed shape are affected by errors with a peak of ~200 mm at the tip of the fiber, that is, 13% of the total length of the sensorized fiber (see (*55*)).

Motion-capture data are acquired using an eight-camera Vicon system (Vicon Motion Systems Ltd., UK) operating under controlled lighting and background conditions to ensure optimal marker visibility. The system is connected via a network switch to a dedicated desktop workstation for synchronized acquisition. ELEANOR is instrumented with 18 reflective markers distributed from base to tip, arranged along five longitudinal lines corresponding to four body segments. Each section is tracked by at least three markers to enable local reference frame reconstruction, while additional markers are placed to compensate for potential occlusions during large deformations or objects grasping. Depending on whether dorsal or ventral tendons are actuated, markers are positioned on the opposite side of the body to minimize mechanical interference with the motion.

Prior to performing experimental trials, regardless of the aim of the experiment, the robot is placed with its base oriented at 45° with respect to the horizontal plane, and motors are individually pulled stepwise (with small rotations) until the pull produces a visible contraction of the robot. After all the motors are actuated, it is verified that no tendons are loosened, and possible visible asymmetry introduced by this procedure is corrected by additional pulling of one or more tendons. Then, the obtained configuration is assumed as home pose, and trials are performed. In all the trials, the recruited tendons are pulled by the same length (260 mm, corresponding to 30% of the undeformed trunk length) at constant speed for 5 s. Then, after a pause of 10 s, tendons are released in 5 s. The pause is imposed to verify that the structure can maintain its configuration, especially in the experiments that involve the use of an object. During these experiments, objects are manually fed to the robot and allowed to fall upon release of the tendons.

***Objects for experiments on grasping***

Objects are obtained via 3D printing using an Ultimaker S7 (Ultimaker B.V., NL) with a commercial filament. The dimensions of the objects are selected based on behavioral experiments reported in (*32*). In that work, an adult African elephant who had an injured trunk (the tip fingers had been severed) interacted with objects with length ranging from 37.5 mm to 300 mm. Preserving the ratio between the length of the trunk and the size of the object, we used objects with size 75-100 mm, aiming at performing whole-body grasping as done by the injured elephant. Based on these considerations, our small cube and cylinder correspond to the objects with dimension $d_3$ = 150 mm in (*32*). Healthy elephants tend to use their tips to grasp smaller objects. Hence, we did not perform a systematic assessment of grasping capabilities with those objects in our study.

**Acknowledgments**

The authors would like to thank Mariana Tavares Pimenta for testing the interaction with ELEANOR, and the ZooSafari (Fasano, Italy) for allowing close observation of elephants.

**Funding:** This work was funded by the European Union Horizon 2020 research and innovation programme under grant agreement No. 863212 (PROBOSCIS project)

**Competing interests:** Authors declare that they have no competing interests.

**Data and materials availability:** All data are available in the main text or the supplementary materials. Datasets from motion capture and FBGS, and Matlab scripts will be available from Zenodo.

# Supplementary Materials for

## Continuous and large-scale: ELEANOR, the soft architected arm inspired by the elephant trunk

Giovanna A. Naselli *et al.*

*Corresponding author. Email: lucia.beccai@iit.it

**This PDF file includes:**

**Supplementary Text**

Dimensions of the bounding trunk

The outer shell of a trunk belonging to an individual of African elephant (available from a shared repository (*1*)) has been used in this work as a reference for the design of ELEANOR's bounding trunk. The dimensions of the biological trunk (length, height and width of the cross section, approximated as elliptical) were extracted and rescaled to obtain a 0.85 m long trunk. ELEANOR's dimensions are reported in the drawing in **Fig. S1**. The reconstructed shape of the biological trunk is affected by static deflections. First, during the acquisition of its geometry, the trunk rested on a rigid flat surface, which is the reason for the flattening of its ventral region. Second, as the trunk rests horizontally, it undergoes vertical deflection due to its self-weight, which distorts each cross section. Despite these deformations, we could question whether the artificial trunk should have linear or nonlinear tapering. **Fig. S2** shows that, at least for the statically deflected sample, a quadratic law represents a better fit for the profile of the trunk. While we do not claim that such tapering is certainly found in trunks of living elephants, we remark that quadratic tapering is more convenient from a mere structural and robotic perspective.

3D printing of large-scale soft parts

The 3D printer used to build the robot, LC Magna (Photocentric, UK) has a build volume of 510×280×350 mm that allows for printing the robot's body in a single job. In fact, the robot's geometry is divided into three parts and printed with the layout shown in **Fig. S3**. The 3D model is split at transversal planes passing through the center of the lattice nodes to ensure an even and flat surface at each resulting interface. This ensures enough surface area for bonding the three parts together at a later stage.

Initially, it has been necessary to determine the printability of different lattice structures by increasing unit cell sizes. One of the most common issues faced when 3D printing any object is ensuring that all overhangs of 45° or more are adequately supported to avoid failures. Generically, lattices are 'self-supporting,' however this may not be true for very large unit cells or certain cell structures. Some flexible or elastomeric materials may exacerbate this issue due to their tendency to move during the print, due to the behavior of uniform lattices when subjected to the forces of delamination from the vat. This occurs between every printed layer as the platform and model move upwards and peel the solid resin from the vat skin (FEP film, 100 µm). To combat this, it is commonplace to orient the model to reduce the surface area of each layer as much as possible, however for a uniformly latticed part the surface area remains approximately constant depending on the bulk dimension of the model.

As mentioned in the main text, the development of Elastic DL160A® was focused on maximizing tensile strength and Young's modulus of the material in the 'green' state. The measured green modulus was 1.48 MPa, with a green ultimate tensile strength of 0.98 MPa (ASTM 412, Type C).

Assembly of the three printed sections requires a strong adhesive at the two interfaces created when slicing the original monolithic model into three parts. Several adhesives were at first proposed which successfully bonded the sections. However, this introduced two planes at which the mechanical properties differed from the bulk, resulting in mechanical discontinuities. The adhesives themselves were more rigid than the robot material, and their brittleness frequently caused fracture at the bond interface in the assembled robot. For this reason, the resin itself is used to perform gluing.

Homogenized properties of the lattice

To estimate the apparent mechanical properties of the lattice, simulations in SOFA are performed on cubic samples of lattice made of DL160A resin (**Fig. S4**). The samples considered have edge length $e$ = 50 mm, with lattice having unit cell size $a = 25$ mm and strut diameter $d = \{1.5, 2.0, 2.5\}$ mm. Other unit cell sizes were considered and simulated but are not discussed here for brevity.

In SOFA, each cube is simulated under compression and elongation, imposing a displacement on the top face and simulating a rigid support on the opposite face. Since, in operative conditions, contact between the lattice unit cells and the external object can occur with varying friction coefficient, each simulation is performed with two different constraint conditions: with nodes on the bottom and top face being fully constrained, and with nodes allowed to slide frictionless on the support plane (an example is shown in **Fig. S5**). Also, the size of the cubic samples is chosen to reflect the dimensions of the distal region of the robot, which provides the greatest contribution to grasping. The solid geometries are meshed with tetrahedral elements, and the imposed displacement is applied incrementally (1 mm/minute). The material is modelled as linear elastic, with Young's modulus equal to 4 MPa (which corresponds to the 100% modulus reported in the technical data sheet of DL160A resin (*2*)). The output of the simulation is the set of reaction forces, exported together with the set of incremental displacements. Results are then converted to stress–strain curves using the nominal cross-sectional area ($e^2$) and initial height ($e$). Results obtained with the two different boundary conditions are averaged and based on such average the apparent modulus is estimated as the secant modulus at 40% strain (since the lattice undergoes large displacements upon actuation and reaches up to such a high strain).

**Fig. S6** shows the results obtained for the lattice used in ELEANOR's body: the lattice exhibits non-symmetric response about the origin, being stiffer under elongation compared to compression. The difference is non-negligible, not only in terms of moduli, but also in trends. Under compression, the secant modulus ranges from 5.8 kPa to 9.8 kPa depending on the strain at which the modulus is calculated, while the tensile secant modulus ranges from 11.0 kPa to 25.6 kPa (right y-axis in the same figure). Moreover, the lattice exhibits softening under progressive compression, while stiffening under progressive elongation.

Results obtained for the other two considered cases, *i.e.*, diameter equal to 1.5 mm and 2.5 mm, supported the decision to opt for a 2 mm diameter. The former is discarded for multiple reasons. Practically, the lattice with such thin struts is more easily subject to failure due to the wear caused by the tendons. Moreover, such a reduced diameter complicates the printing process of large parts, with high chances of failed prints. Finally, simulations reveal that its apparent compressive modulus is smaller than 2 kPa, resulting in a robotic structure with a very low payload (smaller than the average weight of the 3D printed objects that were used for experiments). Conversely, adopting d = 2.5 mm would generate a significantly stiffer structure. Through simulations, we estimate that a lattice with those dimensions has an effective compressive modulus of ~17 kPa for a slightly greater apparent density then that obtained with d = 2 mm, which corresponds to a much greater $\sqrt{E/\rho}$ (hence, greater natural frequencies). Therefore, d = 2 mm represents a good trade-off considering all aspects related to overall structural properties, fabricability, and durability.

The printed cubic samples are analyzed and tested. Their apparent density is obtained by weighing the sample with a digital scale (resolution 0.1 grams) and calculating the ratio between the mass and the volume of the cube. Compression tests are conducted by means of a tensile testing machine (Z005, Zwick/Roell, DE). Elongation tests are not performed as the same boundary conditions of the compression test are hard to replicate; therefore, elongation is studied in simulation only; results are reported in **Fig. S6**.

Oscillations: model-based comparison between ELEANOR and the biological trunk

This section describes the simple mathematical model that supports the results related to the natural frequencies of trunks. The model is used for both the biological and the artificial trunk and is based on the assumptions listed in the main text. As explained there, each trunk is discretized into a series of beam elements, each having two nodes. Each element is characterized by the stiffness matrix that we used in a previous work (*3*), and the following mass matrix (as reported in lecture notes and handbooks, see for example (*4*)):

$$\mathbf{M}_{loc} = \frac{\rho AL}{420}\begin{bmatrix} 140 & 0 & 0 & 70 & 0 & 0 \\ 0 & 156 & 22L & 0 & 54 & -13L \\ 0 & 22L & 4L^2 & 0 & 13L & -3L^2 \\ 70 & 0 & 0 & 140 & 0 & 0 \\ 0 & 54 & 13L & 0 & 156 & -22L \\ 0 & -13L & -3L^2 & 0 & -22L & 4L^2 \end{bmatrix}$$

The matrix above is written for the element in its local coordinates (see (*3*)), with the same system of coordinates (that is usually adopted for beam elements).

As described in (*3*), the stiffness matrix of each element is transformed by means of a rotation matrix, and all the matrices are assembled to generate the stiffness matrix of the entire trunk in global coordinates. The origin of the global system of coordinates is set at the center point of the trunk's base, with $x$ and $y$ axis oriented horizontally and vertically, as shown in **Fig. S7A**.

Once the global matrices have been obtained and constraints have been imposed (the first node of the structure is clamped), it is possible to solve the equations of motion in matrix form:

$$\mathbf{M}\ddot{\mathbf{z}} + \mathbf{K}\mathbf{z} = \boldsymbol{F_g}$$

where **M** and **K** denote the global mass and stiffness matrix of the whole structure, respectively, and $\boldsymbol{F_g}$ models the gravitational force acting on each node (the mass of each element is equally split between the two nodes).

The problem is implemented in Matlab (MATLAB R2023b, The MathWorks, Inc., Natick, MA) and solved by means of the built-in solver *ode45*. Since the vertical ($y$) component of the oscillation is negligible compared to the horizontal ($x$) component, we display the latter (see **Fig. S7B** and **Movie S1**).

Curvature distribution: linear vs. quadratic tapering

Nonlinearly decreasing radius of the cross section helps perform bending and/or curling aimed at effective grasping. Here, a comparison is reported between the planar bending of two arms, which share the same initial and final circular cross sections, but different tapering between them (**Fig. S8**). The arms are modeled as cantilever beams, fully constrained at their largest end (base), and subject to a moment $M$ applied at the free end. The arms differ in how the radius of the cross section decreases from the constrained to the free end. Being $r_0$ the radius of the base, and $r_L$ the radius of the free end, with $r_0 < r_L$, the two cases here considered are:

- a linearly decreasing radius, that is, $r(x) = r_0 - cx$, with $c = (r_0 - r_L)/L$;
- a quadratically decreasing tapering, that is, $r(x) = r_M + ax^2 - bx$, with $a, b > 0$ such that $r(L) = r_m$ and $\left.\frac{dr}{dx}\right|_L = 0$,

Being $L$ the length of the trunk and $x$ the abscissa along the longitudinal axis, with origin ($x = 0$) at the fixed base.

To calculate the deformation, we use the same stiffness matrix used for the dynamic model. In fact, we discretize each trunk as a series of 2-noded beam elements and implement nonlinear matrix structural analysis (*5*) in Matlab environment. All the details concerning the implementation can be found in previous work (*3*).

In the initial configuration, the trunk is unloaded, and its axis is oriented at -45° with respect to the horizontal axis. The first calculation is performed by applying gravitational load incrementally. The mass of each element is proportional to its volume and equally split between the two nodes; hence, self-weight is simulated as a distributed load. The resulting deflected shape is taken as the initial configuration for the second simulation stage, in which the external moment is applied incrementally. **Figure S8** shows the progressive deformation caused by the moment (black lines), starting from the gravitational deflection, for different cases.

As observed in ELEANOR, the structure with quadratic tapering exploits its distal region to create a loop-like shape in its distal region, suitable for wrapping an external object. **Fig. S8A** and **B** show the upward bending of the trunk with quadratic and linear tapering, respectively, generated by $M$. The obtained curvature in the latter is hardly suitable for performing grasping tasks. Increasing the applied moment (i.e., from $M$ to $2M$) results, obviously, in a greater curvature (see **Fig. S8C**), yet the obtained overall deformation of the trunk is not as convenient as in the case of trunk with quadratic tapering. The same holds in the case of downward bending (see **Fig. S8D**, **E** and **F**).

Despite being trivial from the structural viewpoint, this result stresses that trunk-inspired robots can largely benefit from a certain tapering to enhance their grasping capabilities. Moreover, tapering influences not only the stiffness distribution in the trunk, but also the mass distribution, which plays an important role in the dynamics of the trunk.

It is important to note that the artificial trunk is treated here as a homogeneous and isotropic beam, while the biological trunk is composed of muscular and connective tissues, arranged in various orientations, plus the enveloping skin, and therefore is non-homogeneous and non-isotropic.

<u>Deviations from nominal dimensions</u>

All the struts in the lattice adopted in this work should have a diameter of 2 mm, as prescribed by design when the tessellation is generated through 4D_Additive, but we observed that non-negligible deviations from such nominal dimension occur in the robot. Given the large number of struts in the lattice, and the fact that most of them are not easily measurable (as they are far from the outer surface and hence hardly accessible), it is not easy to provide mean value and standard deviation for this measured dimension. However, to provide the reader with a rough estimation, here we report the measurements that we obtained (using a digital caliper) at five sections of the trunk, measuring the diameter of four struts at each section. In the picture in **Fig. S9**, the measured struts are marked with a colored paper label (some of them are hidden by the trunk's body). As the reader can see, it is hard to find regular patterns in such deviations. The thinnest measured strut has a diameter of 1.73 mm, while the thickest of 2.21 mm. Although these deviations might seem negligible, they have a notable effect on the local properties of the lattice, since the bending stiffness of the struts is nonlinearly dependent on the diameter.

Fiber Bragg Grating Sensor

FBGS was encapsulated in a Teflon tube and deployed along the centerline of the robot. The Teflon tube was used to avoid potentially damaging contact between the sensor fiber and the lattice, since the fiber is a sharp nitinol wire with a small diameter (0.3 mm).
FBGS was used for shape reconstruction: the coordinates of 1481 points along the sensorized fiber were streamed and recorded during each trial, together with the curvature along the fiber. This allowed tracking of the movements of the centerline of the robot.
To prevent damaging the sensor due to possible kinking, the tip of the fiber was not constrained to the top plate of the robot. Being free to slide during the experiment, a portion of the fiber protruded out the trunk with variable length.

Motion capture system

*Experimental setup*

The Vicon setup (Vicon Motion Systems Ltd., UK) comprised eight Vero V2.2 infrared cameras arranged around the robot workspace to ensure complete coverage and minimize marker occlusions during large, out-of-plane deformations. A wand calibration procedure was performed before each experimental session, yielding an average residual error below 0.1 mm.
The robotic trunk was equipped with 18 reflective markers distributed from base to tip along five longitudinal lines, defining four main segments. Each line was tracked by at least three non-collinear markers to enable local rigid-body reconstruction, while additional markers were placed to ensure redundancy in case of occlusion. Marker placement was adapted to the specific actuation pattern: for dorsal actuation, markers were placed ventrally, and vice versa, to avoid interference with the moving tendons and ensure continuous optical visibility.
The Vicon system sampled 3D marker positions at 100 Hz and streamed data via a network switch to a desktop workstation running the Robot Operating System (ROS). Data from each trial were stored in independent *rosbag* files containing timestamped marker trajectories. In total, 39 recordings were collected across 14 predefined actuation patterns for both free movements and object grasping.

*Experimental protocol*

Each trial followed a standardized procedure to ensure repeatability for direct comparison across actuation modes and grasping conditions:

- initialization: the robot was positioned in a neutral configuration, and all markers were verified for visibility within the Vicon field of view;
- object placement (for grasping trials): the test object was placed at a fixed reference position relative to the robot base;
- actuation: a predefined tendon pattern was commanded following a pulling–holding–releasing cycle lasting 20 s (i.e., 5 s pulling, 10 s holding, 5 s releasing), performed either in free space or while grasping objects;

- data acquisition: marker trajectories were continuously recorded, synchronized via ROS timestamps;
- rest phase: the robot was returned to its rest configuration before initiating the next actuation sequence.

*Data post-processing and analysis*

All motion-capture data were processed in Python 3.11 using custom routines built on *NumPy*, *SciPy*, and *Matplotlib*. Raw marker trajectories were first filtered using a second-order low-pass Butterworth filter with a cutoff frequency of 10 Hz to suppress high-frequency jitter:

$$p_i^f(t) = Butterworth(p_i(t), f_c = 10Hz)$$

where $p_i(t)$ is the 3D position of the (i)-th marker. Missing data due to short occlusions (<100 ms) were interpolated linearly; longer gaps resulted in exclusion of the affected segment from subsequent analysis. Recordings in Fig. 3E, G, H, and I, originally contained missing data at the beginning of the trials, due to delays between motor control activation and Vicon system initialization. To address this issue, initial pulling phase was reconstructed using backward linear interpolation on deformation modes, based on the values from the first available second of recorded data.

For each of the five longitudinal marker lines, a local reference frame $g_j \in SE(3)$ was defined at each robot section $j$:

- the origin of $g_j$ was placed at the barycenter of the markers defining the corresponding cross-section;
- the normal vector of the section coincided with the tangent to the robot backbone;
- the tangent vector to the section was obtained as the intersection between the local section plane and the global coronal plane;
- the binormal was defined to ensure a right-handed orthonormal triad.

The ordered set of frames $\{g_0, g_1, \dots, g_N\}$ described the discrete backbone configuration along the robot.

The relative transformation between consecutive frames was expressed using the logarithmic map of the special Euclidean group $SE(3)$:

$$\hat{\xi}_i = \frac{1}{l_{r_i}} \, log_{SE(3)}\left(\text{inv}_{SE(3)}(g_{i-1}) \cdot g_i\right)$$

where $l_{r_i}$ is the nominal (rest) length of segment $i$, and $\hat{\xi}_i \in \text{SE}(3)$ encodes the local twist vector representing rotation and translation between consecutive frames.

From $\hat{\xi}_i = [\, \tau_i, \kappa_{t,i}, \kappa_{b,i}, \epsilon_i \,]^\mathsf{T}$, the deformation quantities were obtained as:

- torsion (rotation around the normal direction): $\bar{\tau}_i = \hat{\xi}_i[0]$
- curvature (magnitude of bending in tangential and binormal directions): $\bar{\kappa}_i = \sqrt{\hat{\xi}_i[1]^2 + \hat{\xi}_i[2]^2}$
- axial strain (compression), relative to the nominal segment length: $l_i = l_{r_i} \, \hat{\xi}_i[3]$

with $l_{r_i} = [0.225, 0.225, 0.2125, 0.175]$ m.

These quantities were computed per segment and visualized as spatiotemporal maps of $\kappa(s,t)$, $\tau(s,t)$, and $\epsilon(s,t)$ along the normalized abscissa $s \in [0,1]$. Each trial's heatmaps highlighted the dynamic evolution of bending, twisting, and axial deformation across actuation phases.

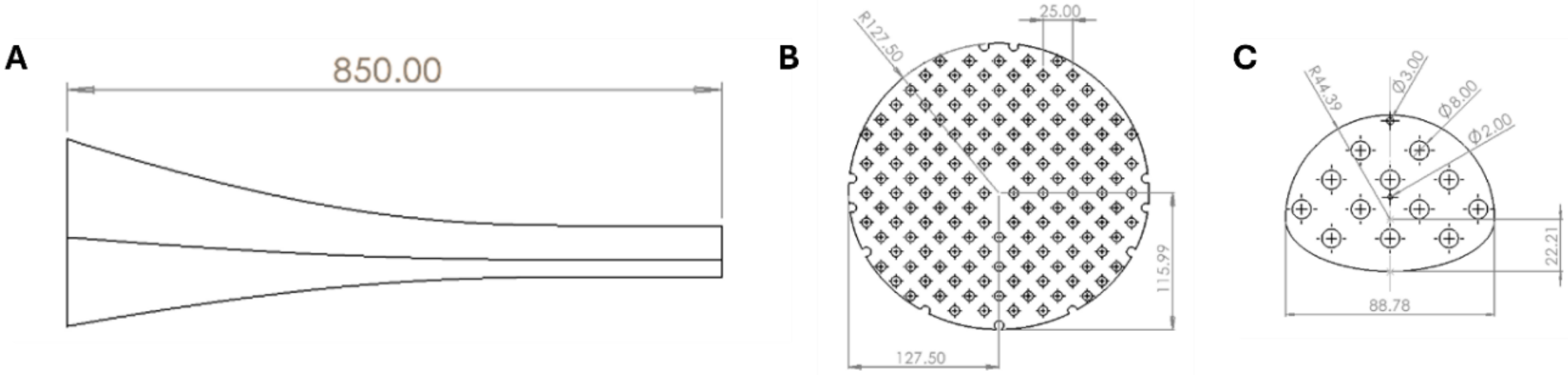

**Fig. S1. Drawings of ELEANOR's outer shell.**

(A) Side view of the outer trunk, showing the total length including the base and the top plate (each 10 mm thick). Views of the base (B) and the distal (C) plate of the robot, respectively. Both plates are 3D-printed with holes (diameter 8 mm) for tendons' routing. Holes are aligned with the center of the unit cells of the lattice. All the dimensions are in millimeters. Views are not shown at the same scale.

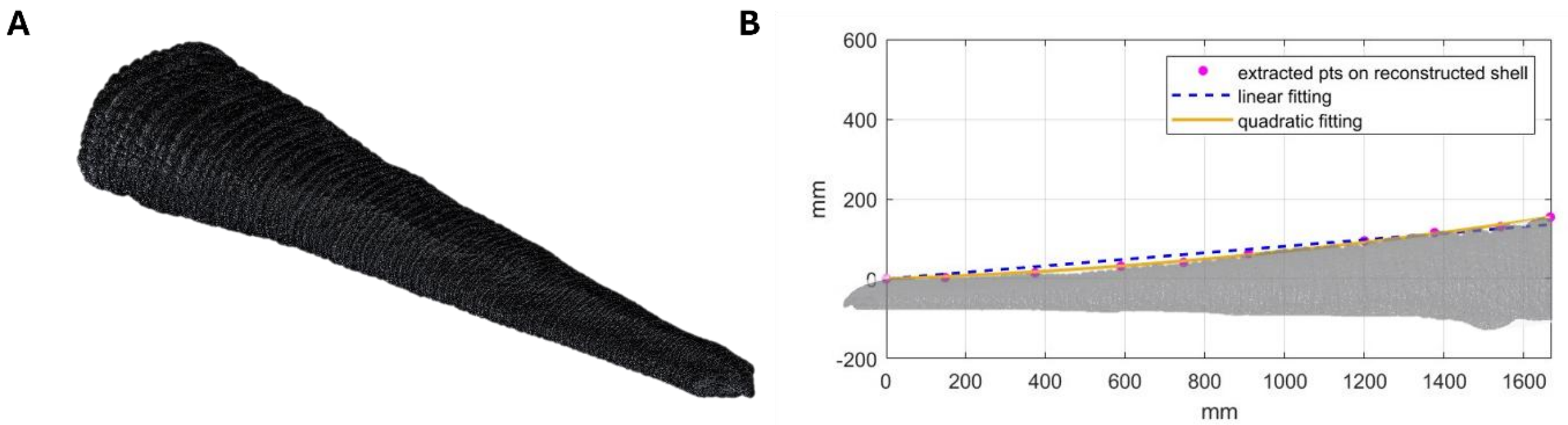

**Fig. S2. Fitting of the dorsal profile of the outer shell of an African elephant's trunk.**

(A) Geometry of the outer shell of an individual of L. africana, available from a public repository (*1*). (B) Side view of the outer shell overlapped with a fitting of its dorsal profile, showing that a quadratic tapering better captures such profile compared to a linear fitting. However, it must be considered that the shell was obtained for a trunk resting on a rigid surface, which is the reason for its ventral side to be flattened. The trunk is also deformed by its self-weight. Therefore, we do not have strong evidence that elephant trunks are characterized by nonlinear tapering, yet the latter allowed for bending and curling useful for whole-body grasping.

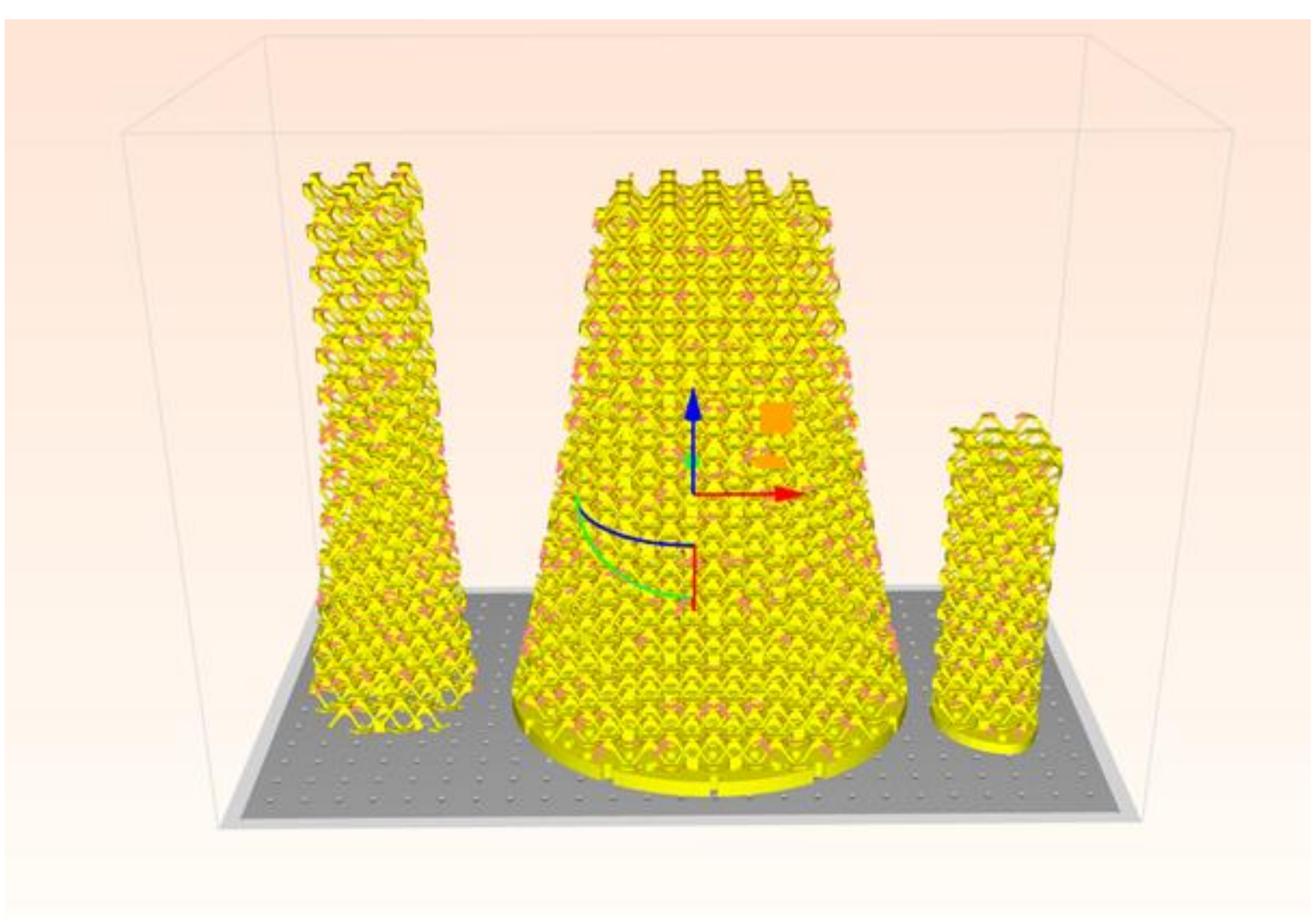

**Fig. S3. Rendering of the 3D printed parts in the build volume of the printer.**

The trunk's body was obtained by glueing the three parts shown in the figure. Except for the plates of the proximal and the distal part (in the center and on the right, respectively), the other end sections consist of nodes split in halves, to facilitate glueing after the printing.

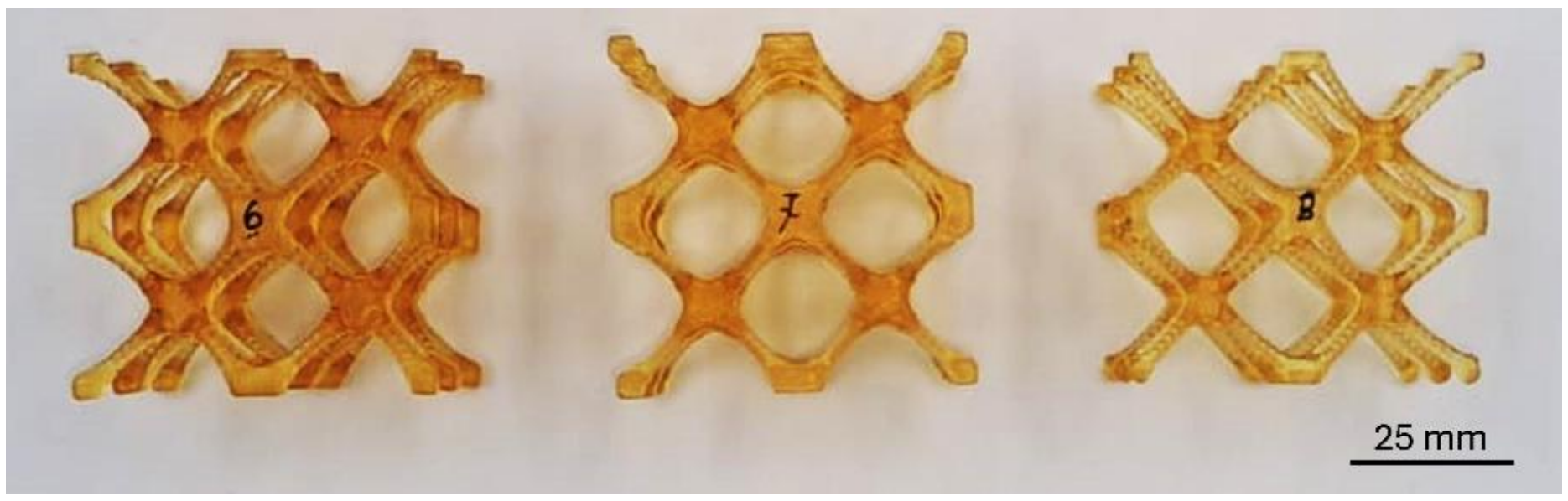

**Fig. S4. 3D printed cubic samples.**

Lattice samples for assessing the quality of prints and for compression tests. Unit cell is 25 mm, strut diameter is equal to 2.5 mm (left), 2.0 mm (center), or 1.5 mm (right).

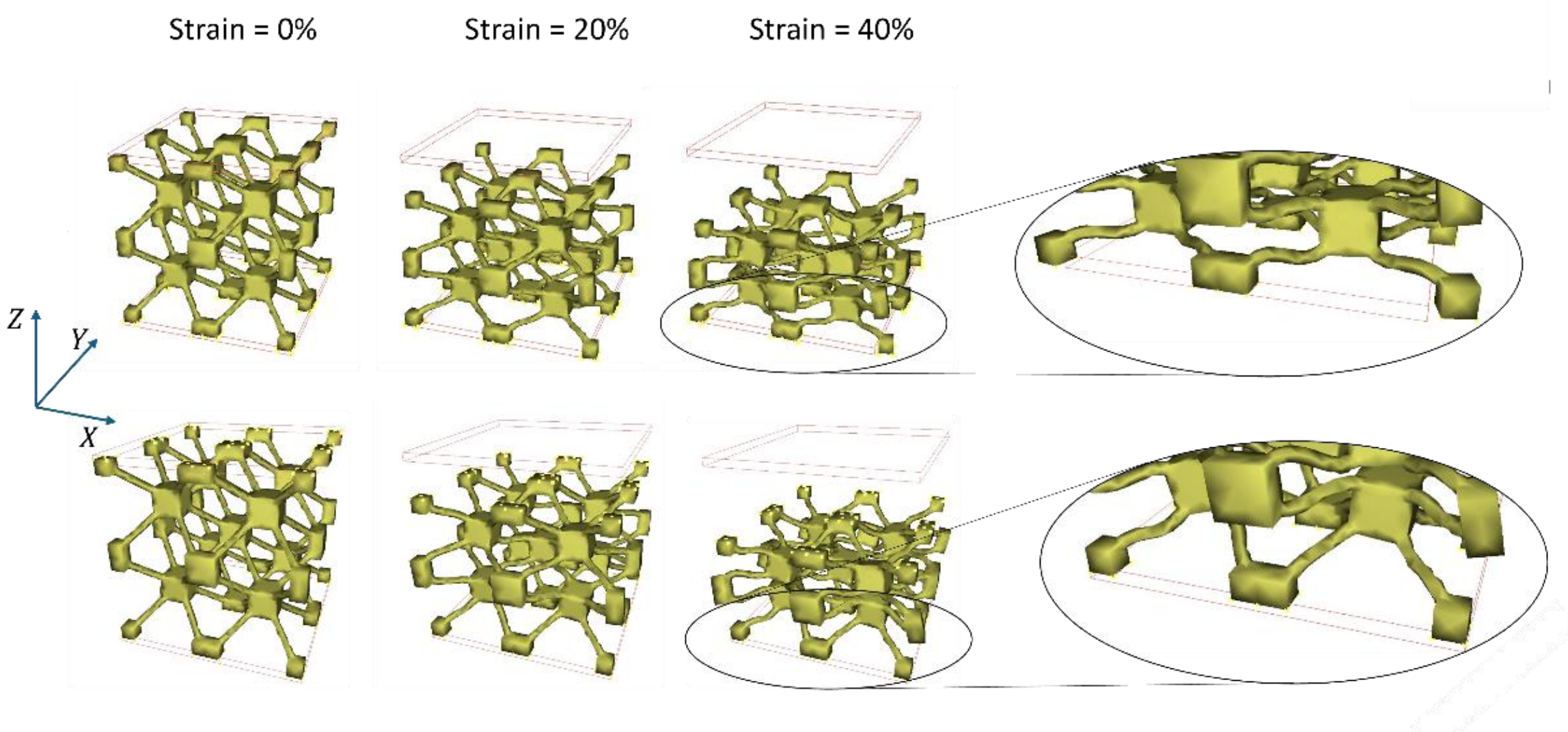

**Fig. S5. Example of simulated compression of a lattice cubic sample.**

Compression is simulated by constraining one face of the cube and imposing an increasing displacement along the Z-axis at the opposite face. Two different conditions are simulated: frictionless (top), in which the nodes located at the top and bottom faces are allowed to slide freely on the XY-plane and to rotate, and clamped (bottom), meaning that the nodes at the bottom and the top face are fully constrained to the plane containing the face.

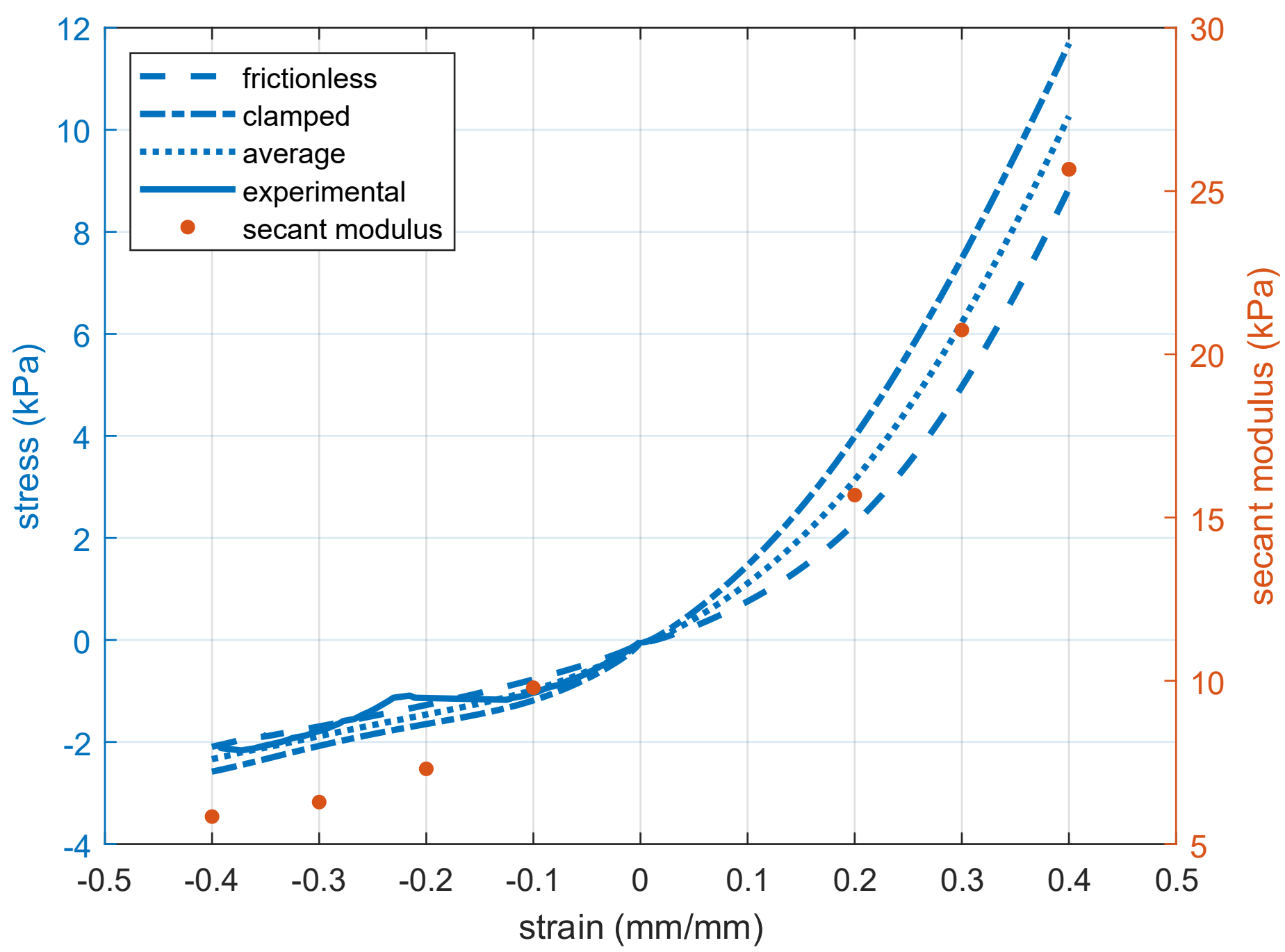

**Fig. S6. Simulation and experimental results of a lattice-based cubic sample with e = 50 mm, a = 25 mm and d = 2 mm.**

Results are shown for both compression and elongation (corresponding to negative and positive strain, respectively), and for both the constraint conditions (clamped and frictionless). Results obtained with the two constraint conditions are averaged (dotted line) and used to estimate the apparent elastic modulus of the sample. For validation against experimental data, compression tests are performed on a 3D printed cubic sample with the same dimensions (solid line, for negative strains).

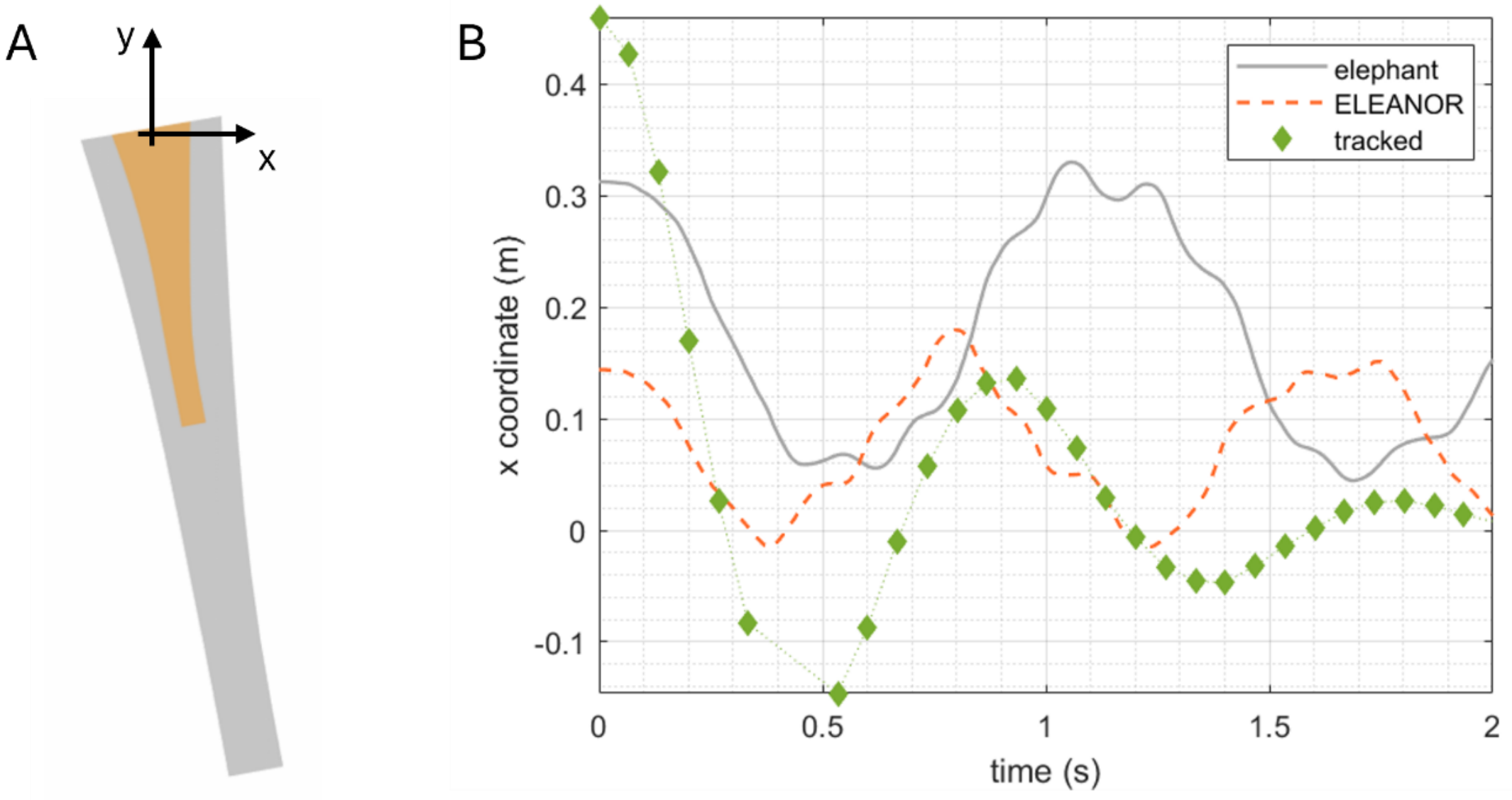

**Fig. S7. Oscillations of ELEANOR vs. real trunk.**

(A) Both the biological (gray) and the artificial (orange) trunk are modelled (separately, although here their overlapping is shown) in the denoted xy-plane. (B) Graph showing the horizontal component of oscillations of natural (gray) vs. artificial (orange) trunk over time, as obtained by modeling. The green dotted graph shows the experimental result of ELEANOR oscillation.

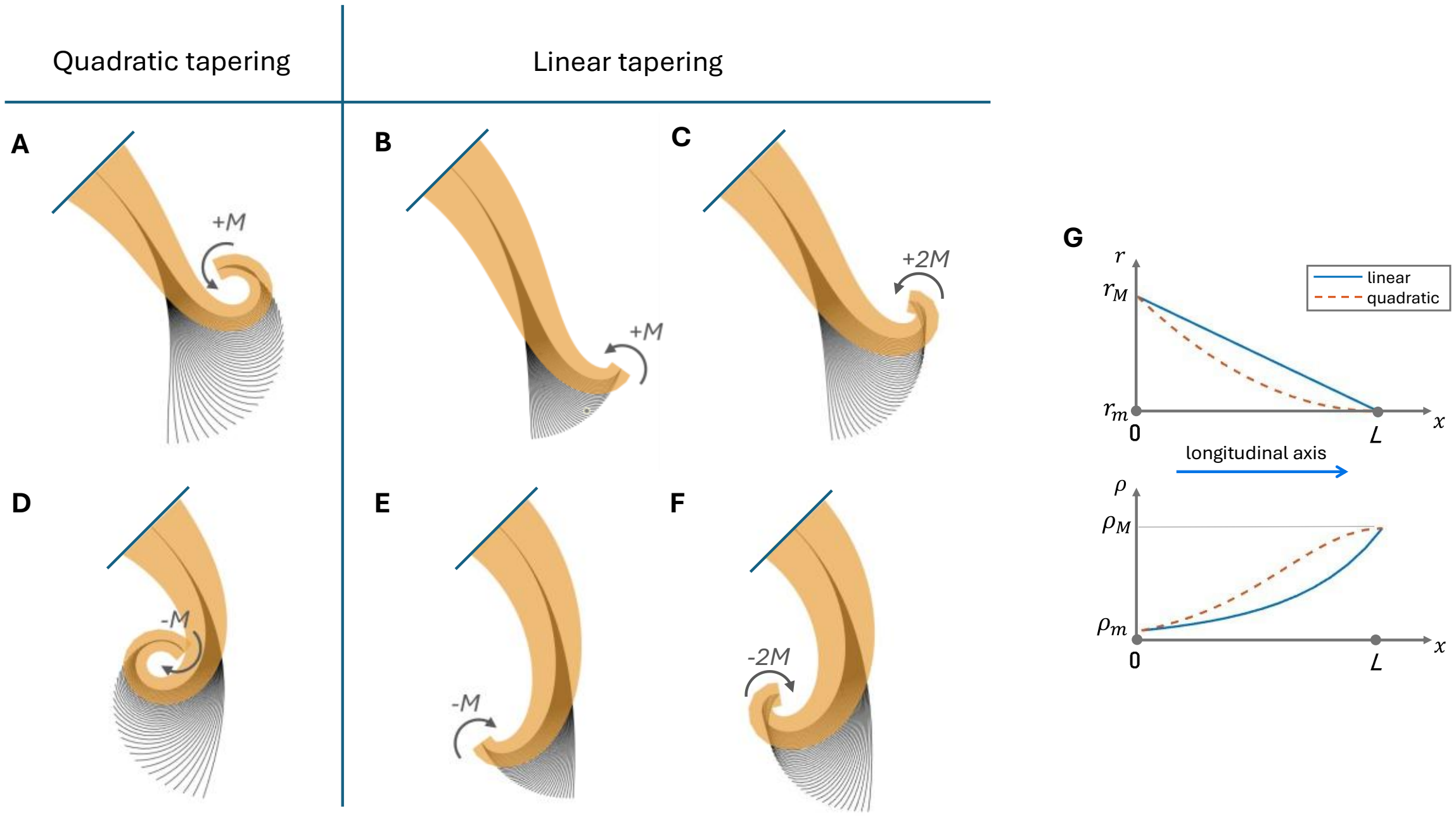

**Fig. S8. Planar bending for two trunk-like bodies with different longitudinal tapering.**

The positive moment M, counterclockwise, generates dorsal (*i.e.*, upward) bending of the trunks (first row), inducing a curvature distribution that is suitable for grasping in case of quadratic tapering of the structure (**A**). The same moment is not equally effective for grasping with the trunk having linear tapering (**B**). Increasing the applied moment (from *M* to *2M*, hence increasing the mechanical work for actuation) to generate a larger curvature does not yet result in a deformed configuration that is as convenient as the one in the first case (**C**). Obviously, the same holds in case of clockwise moment to generate ventral (*i.e.*, downward) bending, as shown in the second row (**D**, **E**, **F**). (**G**) For the shown examples, the simulated structures have same maximum, proximal radius $\boldsymbol{r_M}$ and minimum distal radius $\boldsymbol{r_m}$. Only their tapering determines the different distribution of curvature $\boldsymbol{\rho}$ along the length.

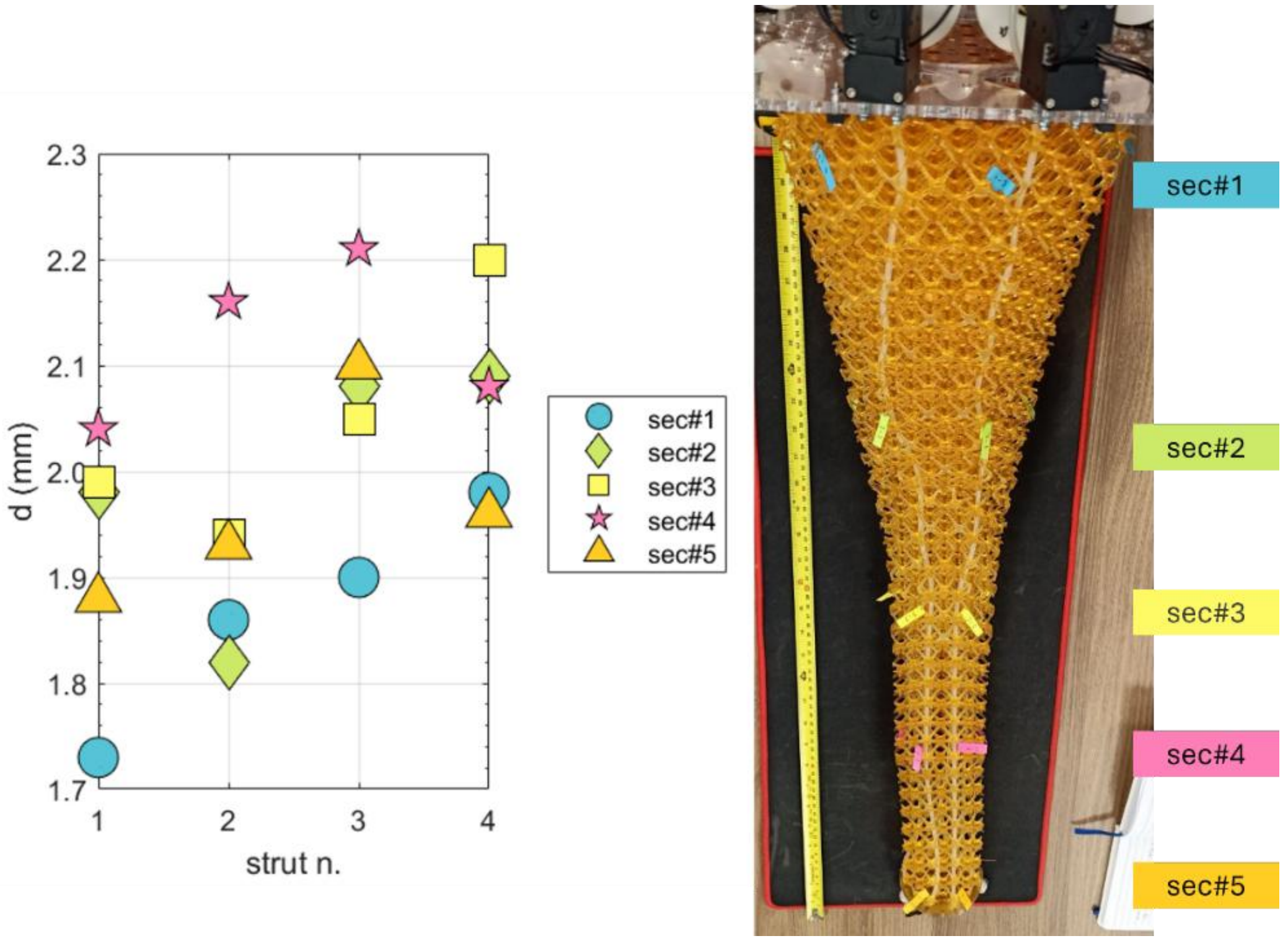

**Fig. S9. Measured deviations from nominal diameter of the lattice struts.**

The plot reports the measured values of the diameter of struts distributed along the robot's body, which differ from the nominal value (2 mm). Five equally spaced sections are considered, randomly selecting four struts at each section. Colored labels in the picture (left) show the measured struts (some are not visible as they are covered by the robot itself). Measurements are obtained using a digital caliper, measuring the thickness at the midpoint of the struts.

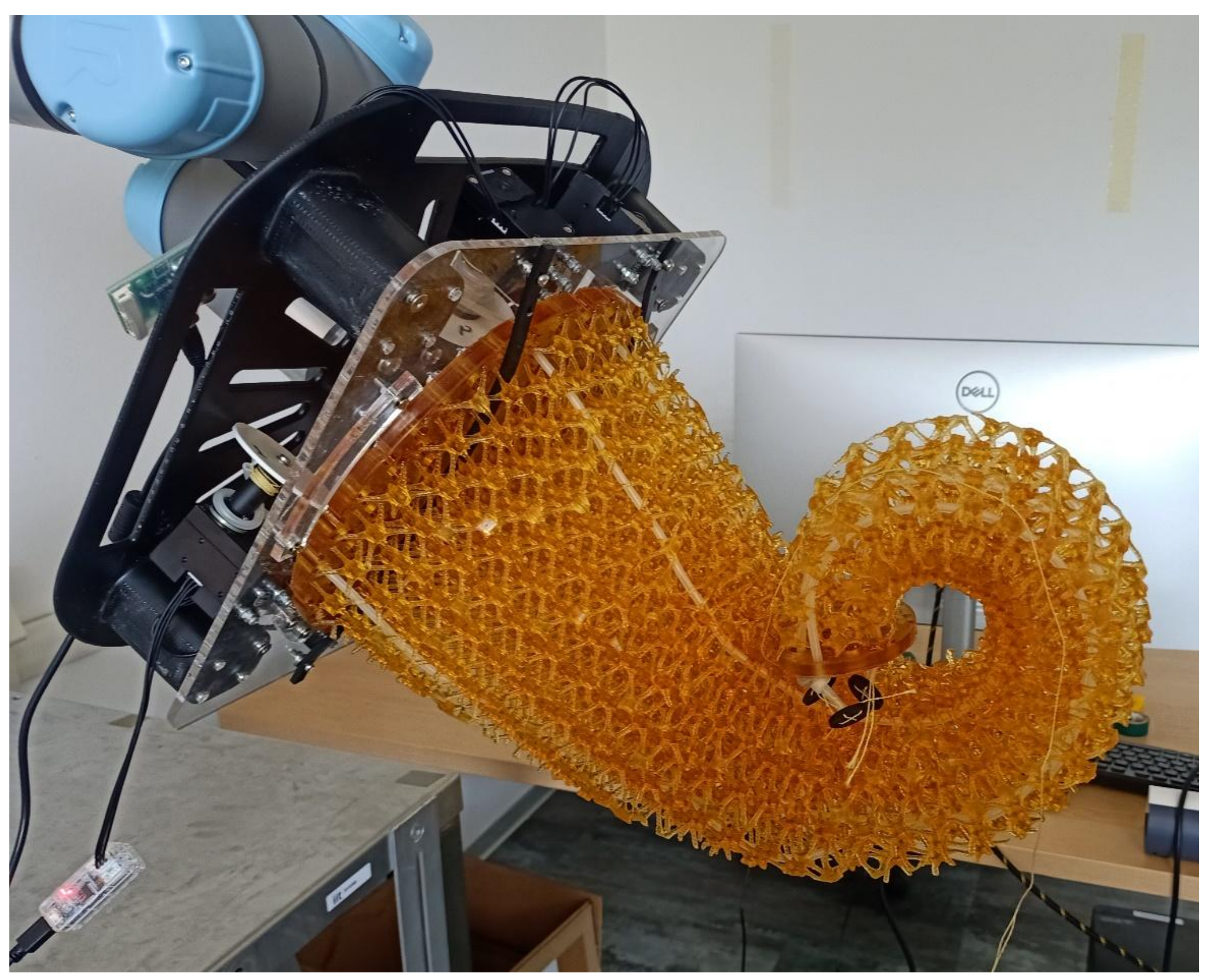

**Fig. S10. Rigid platform connecting ELEANOR to UR5e.**

ELEANOR’s base is fixed to a plexiglass plate to which also the motors are fastened. Once the tendons are wound around 3D-printed pulleys attached to the motors’ axes, the plexiglass plate is fixed to the black frame (obtained through fused filament fabrication) that serves as interface to the UR5e. The platform is sufficiently large to host the motors.

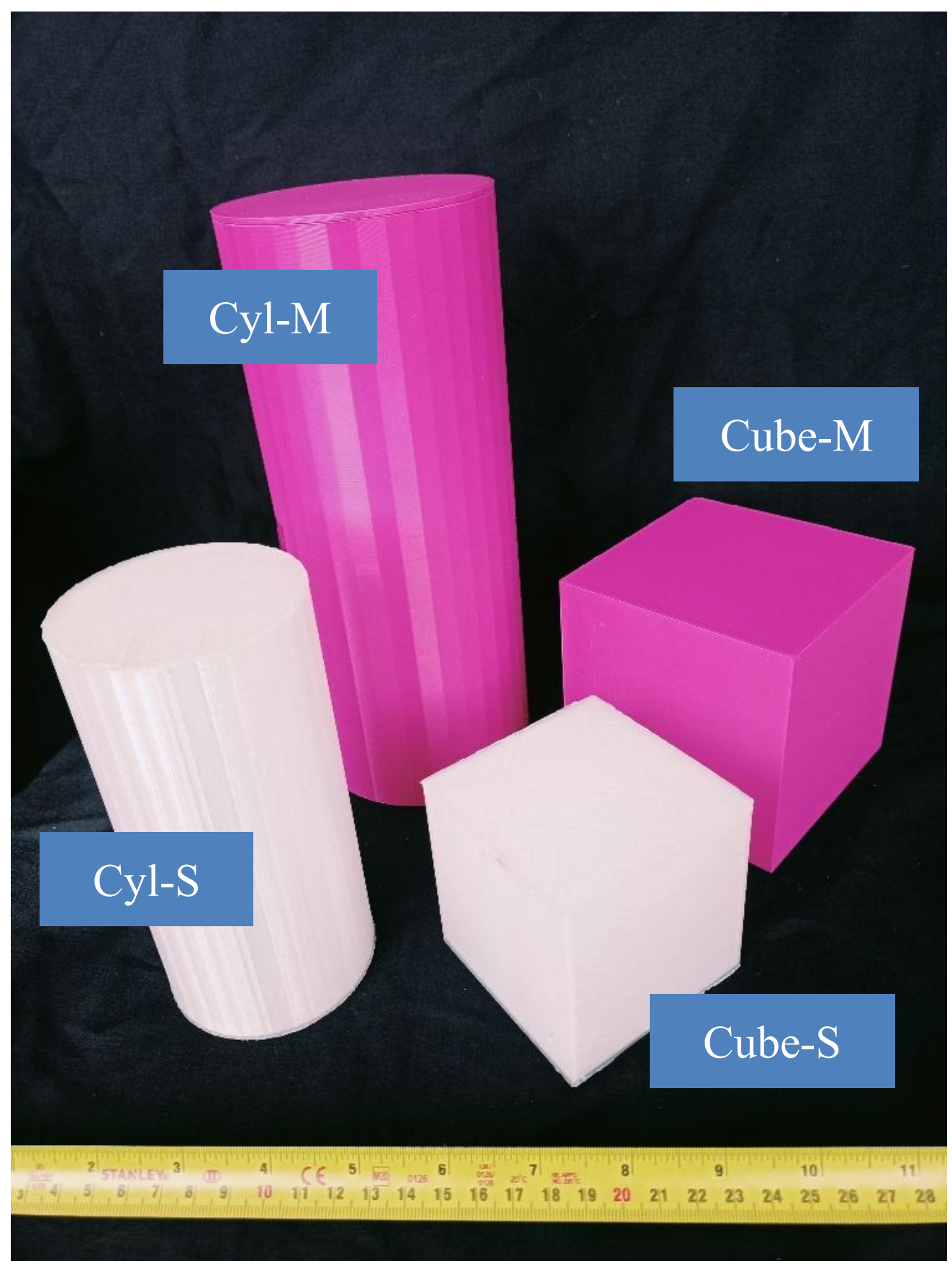

**Fig. S11.Objects used for experiments on grasping.**

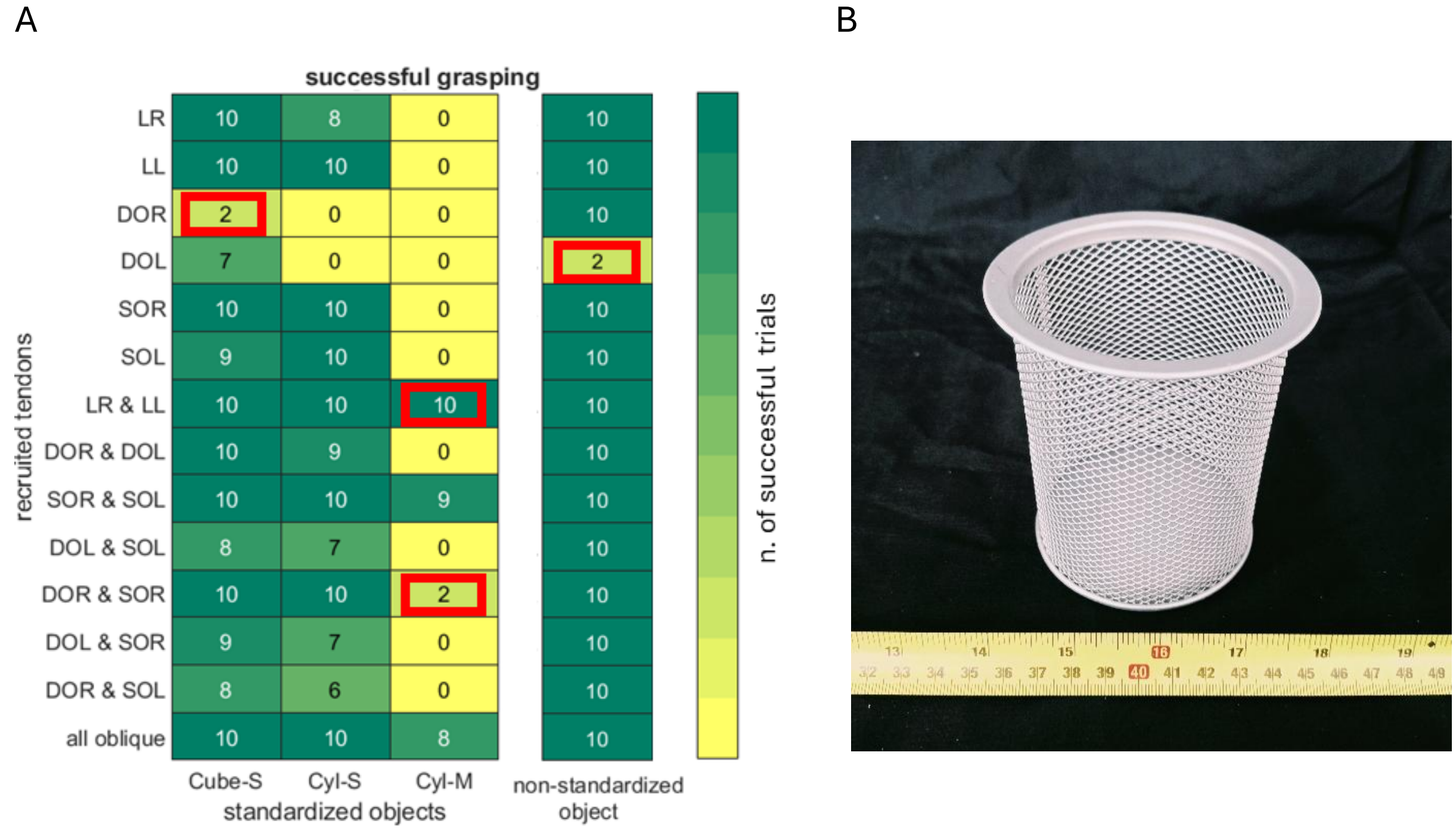

| recruited tendons | Cube-S | Cyl-S | Cyl-M | non-standardized object |
|---|---|---|---|---|
| LR | 10 | 8 | 0 | 10 |
| LL | 10 | 10 | 0 | 10 |
| DOR | 2 | 0 | 0 | 10 |
| DOL | 7 | 0 | 0 | 2 |
| SOR | 10 | 10 | 0 | 10 |
| SOL | 9 | 10 | 0 | 10 |
| LR & LL | 10 | 10 | 10 | 10 |
| DOR & DOL | 10 | 9 | 0 | 10 |
| SOR & SOL | 10 | 10 | 9 | 10 |
| DOL & SOL | 8 | 7 | 0 | 10 |
| DOR & SOR | 10 | 10 | 2 | 10 |
| DOL & SOR | 9 | 7 | 0 | 10 |
| DOR & SOL | 8 | 6 | 0 | 10 |
| all oblique | 10 | 10 | 8 | 10 |

**Fig. S12. Results from whole-body grasping experiments.**

(A) Number of successful grasping trials out of ten, in experiments performed with four objects. The non-standardized object of the single column on the right is the metallic penholder shown in (B).

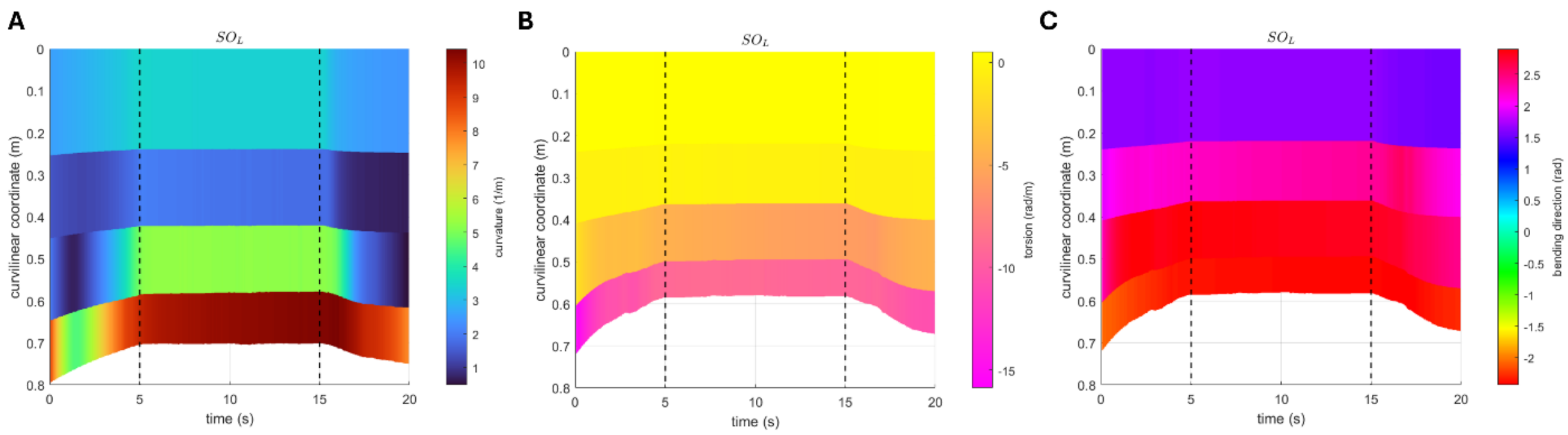

**Fig. S13. Kinematics during grasping with out-of-plane deformation.**

Results obtained by means of motion capture in a trial in which ELEANOR grasped Cyl-S upon use of the SOL tendon. (A) Curvature, (B) torsion and (C) bending direction. Compared to free motion, the rotation of the bending plane localizes in the medial/distal segments, and torsion increases near the payload, yielding a more helical distal configuration.

**Table S1. Properties used for dynamic simulation.**

| | density | elastic modulus | radius at base end | radius at top end | length |
|---|---|---|---|---|---|
| **ELEANOR** | 90 $kg/m^3$ | 15 kPa | 118 mm | 33 mm | 830 mm |
| **biological trunk** | 1180 $kg/m^3$ * | 938 kPa ** | 210 mm | 75 mm | 1800 mm |

* See reference (*6*) of the main text
** See reference (*7*) of the main text

**Table S2. List and dimensions of objects used for evaluation of grasping performance.**
Four cylindrical objects and two cubic objects were used in experiments aimed at assessing grasping capabilities. Cyl-S, Cyl-M, Cube-S and Cube-M are shown in Fig. S11. Cubes larger than Cube-M were built but not used due to grasping failure with objects similar in size to Cyl-L1 and Cyl-L2.

| Object ID | Description | Dimensions | |
|---|---|---|---|
| | | *diameter (mm)* | *height (mm)* |
| Cyl-S | small cylinder | 75 | 150 |
| Cyl-M | medium-size cylinder | 100 | 240 |
| Cyl-L1 | large cylinder | 150 | 150 |
| Cyl-L2 | large cylinder | 125 | 250 |
| | | *edge length (mm)* | |
| Cube-S | small cube | 75 | |
| Cube-M | medium size cube | 100 | |

**Movie S1.**
The similarity between ELEANOR and the natural trunk for dynamic behavior.

**Movie S2.**
Interaction between a human operator and ELEANOR.

**Movie S3.**
Highlights of whole body grasping performed by ELEANOR.